\documentclass{article}
\usepackage{iclr2027_conference,times}
\usepackage{enumitem}
\usepackage{amsmath,amssymb}
\usepackage{graphicx,booktabs}
\usepackage[font=small]{caption}
\usepackage{xcolor}
\usepackage{pifont}
\newcommand{\cmark}{\textcolor{green}{\ding{51}}}
\newcommand{\xmark}{\textcolor{red}{\ding{55}}}
\newcommand{\pmark}{\textcolor{orange}{$\triangle$}}
\newcommand{\proposedone}{\textsc{ManiPhysicsZoo}}
\newcommand{\proposedtwo}{\textsc{ManiPhysicsBench}}
\usepackage{wrapfig}

\usepackage{hyperref}
\usepackage{url}

\title{\proposedtwo: Physics-Based Assessment of Object Preservation in VLA Manipulation}
\newcommand{\mail}[1]{\href{mailto:#1}{\includegraphics[width=0.23cm]{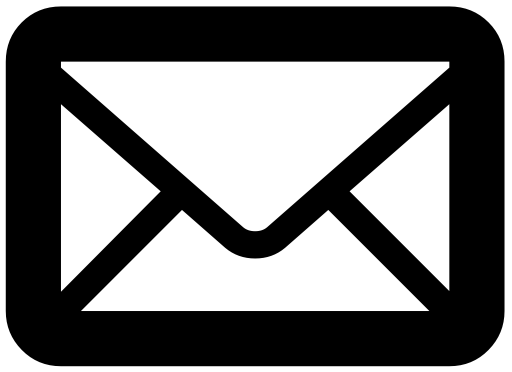}}}
\author{Sangwu Park$^{1, \mail{sangwu.park@kaist.ac.kr}}$, Yeonjun In$^{1, \mail{yeonjun.in@kaist.ac.kr}}$, Wonjoong Kim$^{1, \mail{wjkim@kaist.ac.kr}}$, \\
\textbf{Sungwon Kim$^{1, \mail{swkim@kaist.ac.kr}}$, Sein Kim$^{1, \mail{rlatpdlsgns@kaist.ac.kr}}$, Chanyoung Park$^{1, \mail{cy.park@kaist.ac.kr}}$\thanks{Corresponding Author}} \\
$^1$KAIST
}

\iclrfinalcopy

\begin{document}
\maketitle

\begin{abstract}
Vision-language-action (VLA) models aim to perform diverse manipulation tasks, but task success in existing rigid-body benchmarks does not indicate whether they preserve objects. We introduce \proposedone{}, which consolidates literature-supported material properties, 3D meshes, and supporting references into reusable object assets. Using these assets, a solver-based assessment computes grasp-specific damage thresholds from object geometry, material properties, and recorded grasp conditions and compares them with recorded contact forces to assess potential deformation and fracture. Building on these components, \proposedtwo{} evaluates object preservation in LIBERO and SimplerEnv across three physics axes and three difficulty levels. Public VLA checkpoints show a substantial gap between task success and safe success, defined as task completion while preserving the object. Their gripper commands concentrate near full opening and closure, with largely similar aggregate distributions across objects, consistent with binary gripper supervision. We examine how object-specific continuous gripper labels change model behavior by retraining a VLA model. The retrained model shows more object-dependent gripping and higher safe success, but lower task success and limited generalization of object-preserving behavior.

\end{abstract}

\section{Introduction}
\label{sec:introduction}
\vspace{-1ex}
Vision-language-action (VLA) models aim to perform a wide range of manipulation tasks from visual observations and language instructions. Their recent performance and generality across tasks and embodiments have made them a major paradigm in robot learning~\citep{rt2, openvla, pi0, gr00t}. This capability has been evaluated through manipulation tasks in simulation and in the real world, which serve as a standard protocol for VLA evaluation~\citep{libero, simplerenv}. 

However, task completion alone does not establish whether a policy handles objects without damaging them. In real-world deployment, VLAs may need to grasp and move objects with diverse physical properties, such as fragile eggs and rigid wooden boxes. Humans take these properties into account by holding fragile objects firmly enough to prevent slipping but gently enough to avoid damage, while using a stronger grip when handling rigid objects. In this sense, physical properties determine which actions are appropriate, and a genuinely generalist policy should be able to complete a task while preserving the object. We therefore ask: \textbf{Can current VLAs account for objects’ physical properties to complete manipulation tasks without causing damage?}

The direct way to evaluate this ability is to conduct real-world experiments. However, fractured or permanently deformed objects cannot be reused under the same conditions, so repeated trials incur object replacement costs. Moreover, damaged objects and fragments may pose safety risks to the robot and its surroundings. Therefore, a simulation-based framework is needed to assess object preservation under repeatable conditions. Yet widely used simulation benchmarks do not account for object damage when evaluating task success~\citep{libero, simplerenv}. Because objects are modeled as rigid bodies, a policy can receive the same success score regardless of whether it handles the object carefully or uses gripping that would damage it in the real world. Adjacent lines of work, including physical understanding benchmarks, robot safety benchmarks, and tactile and force-aware VLAs, leave this gap open, as none of them judges a policy's actions against a physically grounded criterion for object preservation~\citep{physbench, manipbench, safevla_bench, oopsieverse, tactile_vla, taf_vla}. These limitations highlight the need for a common evaluation protocol that assesses whether policies complete tasks while respecting each object's physical constraints.

Developing such an evaluation protocol requires defining preservation criteria for each object and reliably assessing how policy-applied forces affect its state during a rollout. This entails two challenges. First, defining physically grounded preservation criteria requires object-specific material properties, geometric meshes, and supporting references, but these resources are not readily available in a unified, verifiable form. Second, evaluating object preservation requires translating the forces applied during a rollout into predictions of deformation or fracture, but the rigid-body simulators used in standard manipulation benchmarks do not model these outcomes.

To address the first challenge, we propose \proposedone{}, an object library that consolidates object-specific material properties, 3D meshes, supporting references, structural information, and validation status into reusable assets. These assets can be reused for evaluating new policies, robots, and objects across simulation environments. The supporting references justify the material properties and preservation criteria, making their physical basis traceable and verifiable.
 
To address the second challenge, we introduce an assessment method that uses physics solvers to determine whether contact forces recorded during a manipulation rollout could cause object deformation or fracture under the corresponding grasp conditions.
After each rollout is completed, the physics solver combines the recorded grasp information with the material properties and 3D mesh from \proposedone{} to compute a damage threshold for each grasp. We compare the recorded contact forces with the corresponding thresholds and judge an interaction to be potentially damaging if its force exceeds the threshold. We justify each threshold using material properties and damage criteria supported by the literature, and document the modeling assumptions to make the resulting evaluation traceable and verifiable.

Building on these two components, we construct \proposedtwo{}, a physics-based benchmark for evaluating object preservation during manipulation. \proposedtwo{} constructs manipulation tasks in LIBERO~\citep{libero} and SimplerEnv~\citep{simplerenv} using objects selected from our \proposedone{}. 

Solver-based assessment enables object-preservation evaluation on these platforms, even though their simulators model objects as rigid bodies. The benchmark is organized into three physics axes: brittle fracture, soft object deformation, and thin-walled container deformation. Each axis includes high, medium, and low difficulty levels, with higher difficulty corresponding to a narrower range of gripping forces that allow the object to be lifted without damage.

We evaluate public VLA checkpoints~\citep{openvla_oft, cogact, pi05, molmoact2} on \proposedtwo{} and observe a substantial gap between task success and safe success, defined as task completion while preserving the object. To investigate this gap, we analyze the policies' gripper commands to examine whether they adapt their gripping behavior to objects' physical properties. We find that these commands concentrate near fully open and fully closed states, with largely similar aggregate distributions across objects.

This pattern motivates examining the role of gripper supervision. Standard VLA pipelines often binarize gripper labels into open and close, which may limit the learning of object-dependent gripper control~\citep{bridgedata_v2, openvla}. We therefore replace binary labels with continuous labels derived from object-specific physical constraints and evaluate how retraining with these labels affects gripper behavior and safe success. Retraining yields more object-dependent gripping and higher safe success. However, task success decreases, and the resulting object-preserving behavior shows limited generalization.

Our contributions are summarized as follows.
\begin{itemize}[leftmargin=1.5em, itemsep=0pt]
\vspace{-3pt}
    \item \textbf{\proposedone.} An object library that consolidates object-specific meshes, properties, and supporting references into reusable assets for physically grounded preservation assessment.
    \item \textbf{Physics solver-based assessment.} A solver-based pipeline that computes damage thresholds for each grasp from object geometry, material properties, and grasp conditions. 
    \item \textbf{\proposedtwo.} A benchmark that evaluates across three physics axes and three difficulty levels whether VLA models manipulate objects while respecting their physical constraints.
    \item \textbf{Evaluation and behavioral analysis.} {An evaluation of public VLA checkpoints, reporting the gap between task success and safe success. We analyze gripper behavior and examine the effects of object-specific continuous gripper supervision, identifying improvements in safe success alongside reduced task success and limited generalization of object-preserving behavior.}
\end{itemize}

\section{Related Work}
\label{sec:related_work}
\vspace{-1ex}
\smallskip
\noindent\textbf{Manipulation Benchmarks.}
Manipulation benchmarks evaluate policies across diverse tasks, objects, and environments. LIBERO~\citep{libero} studies knowledge transfer across manipulation tasks, while SimplerEnv~\citep{simplerenv} provides simulation-based evaluation of real world robot policies. CALVIN~\citep{calvin} evaluates language-conditioned task sequences, and RLBench~\citep{rlbench} provides a diverse task suite with demonstrations. Because these benchmarks model objects as rigid bodies and do not track whether an object is damaged, task success alone does not indicate object preservation.

\begin{wrapfigure}{r}{0.52\textwidth}
    \centering
    \vspace{-8pt}
    \captionof{table}{Comparison with representative related work. \textit{Simulation}: policies are evaluated on manipulation tasks in simulation. \textit{Preservation}: damage to the object is judged. \textit{Physics}: the verdict rests on the objects' own physical response, either computed from its material properties with a physics model or observed in real tests. \textit{Library}: object assets carry material properties and their sources. \cmark{} full, \pmark{} partial, \xmark{} none.}
    \label{tab:related_work_comparison}
    \vspace{-4pt}
    \footnotesize
    \setlength{\tabcolsep}{2pt}
    \resizebox{\linewidth}{!}{%
    \begin{tabular}{l c cc c}
        \toprule
        & \multicolumn{1}{c}{Evaluation} & \multicolumn{2}{c}{Verdict} & \multicolumn{1}{c}{Assets} \\        \cmidrule(lr){2-2} \cmidrule(lr){3-4} \cmidrule(lr){5-5}
        & Simulation & Preservation & Physics & Library \\
        \midrule
        LIBERO, SimplerEnv     & \cmark & \xmark & \xmark & \xmark \\
        PhysBench              & \xmark & \xmark & \xmark & \xmark \\
        DefGraspSim            & \xmark & \pmark & \pmark & \pmark \\
        SafeVLA-Bench          & \cmark & \xmark & \xmark & \xmark \\
        OopsieVerse            & \cmark & \cmark & \xmark & \xmark \\
        Tactile-VLA, TaF-VLA   & \xmark & \cmark & \cmark & \xmark \\
        \midrule
        \proposedtwo{} (Ours)  & \cmark & \cmark & \cmark & \cmark \\
        \bottomrule
    \end{tabular}%
    }
    \vspace{-8pt}
\end{wrapfigure}

\vspace{-1ex}
\smallskip
\noindent\textbf{Physical Understanding and Interaction.}
Physical understanding benchmarks evaluate VLMs' knowledge of physical properties and interactions through question answering. PhysBench~\citep{physbench} assesses understanding of object properties and physical dynamics, while ManipBench~\citep{manipbench} evaluates low-level manipulation reasoning through multiple-choice questions on interaction keypoints and action trajectories. We share their interest in connecting physical knowledge to manipulation. However, in these benchmarks' question-answering evaluations, models answer questions without executing manipulation actions, so object preservation during task execution is not directly assessed. Even PhysBench's force test relies on VLM-specified gripping forces, while its simulator treats all objects as rigid bodies and does not simulate grasp-induced damage. Our work complements these evaluations by assessing whether contact forces recorded during policy rollouts exceed physically grounded damage thresholds for each grasp.

Deformable manipulation environments aim to simulate objects that change shape. SoftGym~\citep{softgym} provides tasks with cloth, rope, and fluids, while PlasticineLab~\citep{plasticinelab} focuses on shaping elastoplastic materials. However, deformation is the goal of these tasks, so object preservation is neglected and the damage criterion is absent. DefGraspSim~\citep{defgraspsim} computes the stress and deformation of an object under a grasp with the finite element method (FEM). However, its material parameters are assigned from a fixed set rather than sourced from the object's own material, so the computed response is hypothetical. Moreover, in DefGraspSim a gripper is placed at candidate poses produced by a sampler and closes on the object, without an arm and a manipulation task. Therefore, grasping as part of manipulation falls outside its concern.

\vspace{-1ex}
\smallskip
\noindent\textbf{Robot Safety Benchmarks.}
Robot safety benchmarks assess hazards and harmful interactions during policy execution. SafeVLA-Bench~\citep{safevla_bench} scores rollouts on existing benchmarks against safety constraints specified for the robot, the scene, and how the object is carried. OopsieVerse~\citep{oopsieverse} accumulates mechanical, thermal, and fluid damage as a health value that the authors define for each object. Both share our premise that task completion should be separate from safe completion when evaluating VLAs. However, SafeVLA-Bench centers on the safety of the scenario and does not evaluate object preservation. OopsieVerse evaluates object damage, including preservation, but from damage criteria that are annotated manually rather than derived from the object's properties. The authors describe this model as an approximation adopted in place of accurate physical computation, and it judges preservation regardless of where the object is grasped. The physical rigor of these verdicts therefore cannot be ensured.

\vspace{-1ex}
\smallskip
\noindent\textbf{Force and Tactile-Aware Manipulation.}
Force and tactile-aware methods such as Tactile-VLA, TaF-VLA, and DeliGrasp adapt contact behavior to object properties and feedback during execution, and demonstrate this capability through real robot experiments~\citep{deligrasp, tactile_vla, taf_vla}. These methods share our view that appropriate manipulation requires adapting actions to objects' physical properties and contact conditions. However, their focus is on improving models, and evaluation is conducted on dedicated task protocols and real robot setups, where object preservation is judged by visual inspection after each trial. Because each evaluation runs on real hardware, it is subject to cost and safety constraints, and a common criterion under which existing models could be compared remains outside their scope. 

\vspace{-1ex}
\smallskip
\noindent\textbf{Summary.} As summarized in Table~\ref{tab:related_work_comparison}, \proposedtwo{} brings together reusable object assets, grasp-specific damage assessment using physics solvers, and evaluation of public VLA checkpoints on existing manipulation platforms. Using the object geometry and material properties documented in \proposedone{}, physics solvers compute damage thresholds for the grasp conditions recorded during each rollout. We compare these thresholds with recorded contact forces to assess whether task completion also satisfies the object's preservation criteria. Each judgment therefore depends on both the object's physical properties and how the policy interacts with it. This post-rollout assessment supports a common evaluation protocol while retaining the platforms' existing observation and action interfaces and rigid-body simulators.

\vspace{-1ex}
\section{Preliminaries}
\label{sec:preliminaries}
\vspace{-1ex}
We evaluate whether a manipulation policy completes a task while preserving the manipulated object throughout a simulated rollout. A task instance consists of an object $o$, a scene, and a language instruction on a given manipulation platform. At each time step $t$, a policy $\pi$ receives an observation $x_t$ and produces an action $a_t$. Executing these actions in the platform's rigid-body simulator yields a rollout $\tau$ of length $T$. We represent the grasps extracted from $\tau$ as $\mathcal{G}(\tau) = \{ g_k \}_{k=1}^{K}$ with $g_k = (p_k, d_k, f_k)$, where $\mathcal{G}(\tau)$ denotes the collection of extracted grasps, $g_k$ denotes the $k$-th grasp, and $K$ is the number of grasps, which varies across rollouts.
The platforms we evaluate use parallel-jaw grippers, so we take a grasp to be an interval in which both jaws remain in contact with the object for at least 0.2\,s. The contact position $p_k$ is the pair of points where the two jaws touch the object surface, expressed in the object's coordinate frame, and the contact direction $d_k$ is the unit vector between them, along which the jaws compress the object. The contact force $f_k$ is the force transmitted through this pair of contacts under the position servo control we adopt, and we take it as the largest value sustained over the interval.

We use position servo control for the gripper, which matches the opening and closure commands produced by current VLAs and follows the host platforms' settings~\citep{libero, simplerenv}. This controller drives the gripper toward a target opening and generates contact force when an object resists further closure, so policies modulate gripping force through their closing targets. For mechanical assessment, we assume that the grippers are fitted with standardized silicone pads~\citep{kargar2025softpads, achilli2022fingerpad} and account for the reduction in position-servo force as the pads and object deform. App.~\ref{app:gripper_force} gives the rationale for this choice and the controller settings, App.~\ref{app:gripper_contact_fem} the pad model, and App.~\ref{app:fem_thresholds} how the servo coupling enters the damage threshold.

\begin{figure}[t]
  \centering
  \includegraphics[width=0.9\linewidth]{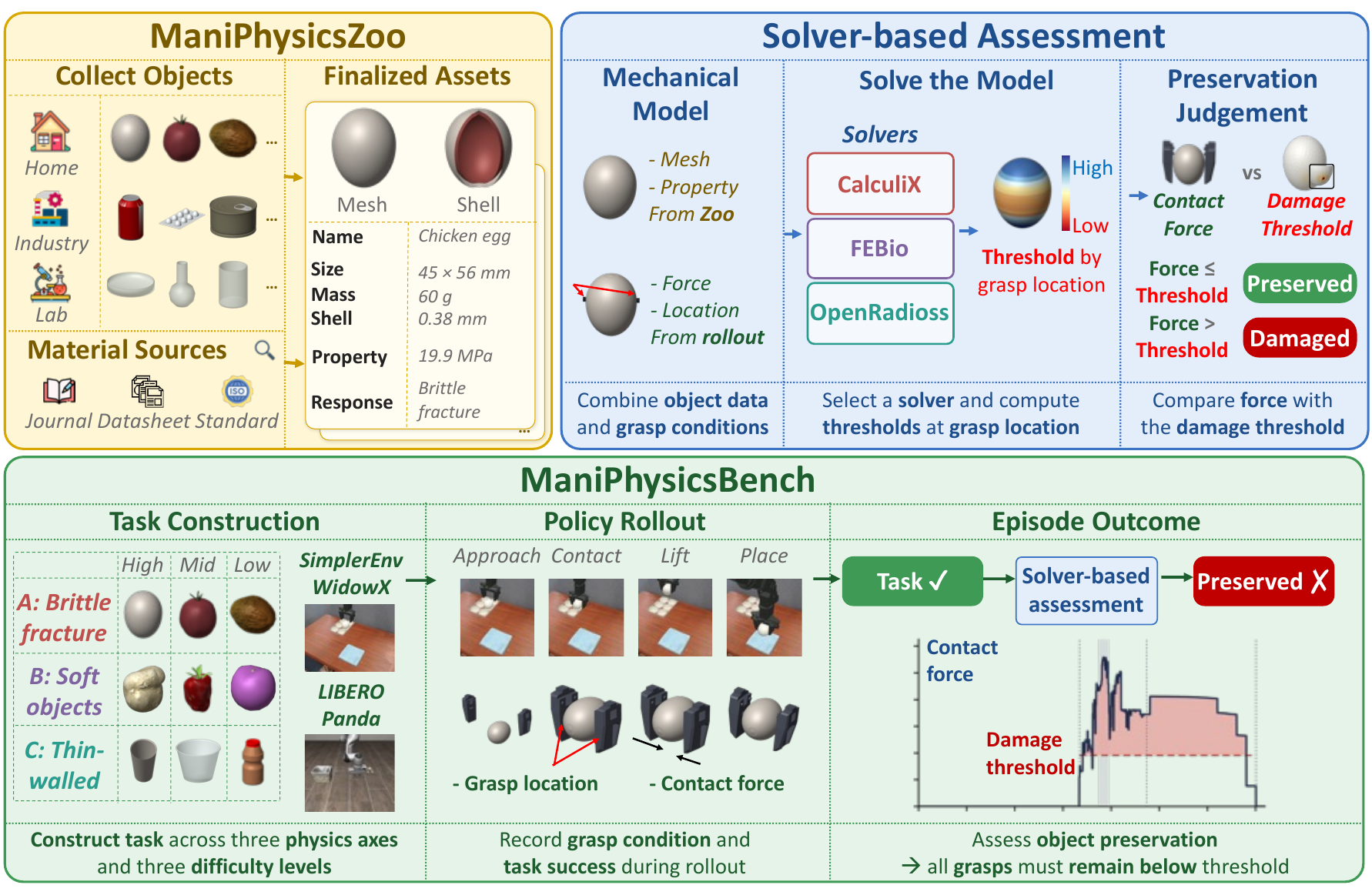}
  \caption{Overview of the framework. \textbf{Upper left:} \proposedone{} turns each object into an asset that carries its geometry, mass, material  properties, and the sources of those properties. \textbf{Upper right:} solver-based assessment builds a mechanical model from the asset and the grasp  recorded during a rollout, computes a damage threshold at that grasp with the solver chosen for the object's response, and compares it with the recorded contact force. \textbf{Bottom:} \proposedtwo{} instantiates tasks on SimplerEnv and LIBERO, and reports whether each episode preserves the object.}
  \label{fig:architecture}
\end{figure}

\vspace{-1ex}
\section{Physically Grounded Preservation Assessment}
\label{sec:assessment}
\vspace{-1ex}
Our framework combines \proposedone{} with solver-based assessment to evaluate object preservation during manipulation. We first describe the construction of the object library, then explain how solver-based assessment uses its assets and the grasp information recorded during policy rollouts to assess potential object damage.
Fig.~\ref{fig:architecture} presents the overall framework. The upper left and upper right panels correspond to \proposedone{} and to solver-based assessment, and the bottom panel shows \proposedtwo{} in Sec.~\ref{sec:benchmark}, which applies both.
\vspace{-1ex}
\subsection{\proposedone}
\label{sec:zoo}
\vspace{-1ex}
We introduce \proposedone{} to provide unified and verifiable object assets for preservation assessment. Because the required meshes and material data are scattered across different sources, we collect and verify these resources for each object. We select objects from household, industrial and laboratory environments according to three criteria. First, the object's size, mass, and graspable region must be compatible with common manipulators, since a task that no robot can perform cannot be evaluated. Second, gripping must be able to cause irreversible damage that compromises the object's intended use. The material properties and structural information needed to assess this damage must be supported by the literature or product standards, providing a traceable physical basis for the damage criterion. This criterion excludes objects such as towels, whose deformation during ordinary gripping does not compromise their intended use. Third, the object must be recognizable in evaluation images, and its appearance must provide cues to the physical properties relevant to gripping. For example, recognizing an egg allows a policy to draw on prior knowledge of its fragile shell, while the visible thin walls of a container suggest that it may deform under gripping forces. These cues provide a basis for adapting gripper actions to the object from visual observations. \proposedone{} comprises 91 objects. The full object list and library details are in Table~\ref{tab:app_zoo_object_list} and App.~\ref{app:zoo}.

We curate the library to cover three types of mechanical response to gripping: brittle fracture, soft object deformation, and thin-walled container deformation. For each object, we first search public 3D mesh collections for a suitable geometry~\citep{objaverse}. Since these meshes are designed primarily for visual representation, they may lack geometric details required for mechanical modeling, such as accurate dimensions or wall thickness. We therefore generate a new mesh when no suitable one is available, add missing structural features such as the wall thickness of hollow objects, and scale the geometry using published product specifications for manufactured objects or reported measurements for natural ones. We visually inspect the resulting meshes for shape, dimensions, and appearance, and retain their physical dimensions, structural information, and mass derived from volume and material density for subsequent mechanical modeling.

With the geometry of each object fixed, we ground the material parameters in reported measurements, material datasheets, or calibration to published mechanical tests rather than choosing them arbitrarily. The objects range from produce to packaging and containers, so their materials include plant tissue, paperboard, plastics, and metals, and their properties are reported in different fields. Therefore we survey established journals in the corresponding field together with material datasheets and product standards for each object. However, these sources do not report the same kind of quantity. For example, a study of eggshell reports a tensile strength~\citep{liu2024eggshell}, a material constant that we can use directly, whereas a study of strawberry reports the force at which the fruit failed under compression at a given loading speed~\citep{an2020strawberry}, an outcome of one test setup with its own contact geometry. We therefore record with each value the part of the object tested, its state, and the conditions of the test, and we determine which physical quantity the value represents. When a source reports a test outcome rather than a material constant, we reproduce that test with the physics solver and adjust the remaining material parameters until the reported outcome is recovered. We then document each parameter as directly reported, calibrated from a reported test, or transferred from a representative material. Where a separate failure test is available, we also check the resulting model against it without refitting; for example, the egg shell model predicts the measured breaking loads of half-eggshell tests within 8--21\% (App.~\ref{app:literature_justification}). 

We represent each object asset as $o = (n, \mathcal{M}, \mathcal{P}, \mathcal{R})$, where $n$ is the object name and $\mathcal{M}$ contains the object's mesh, physical dimensions, mass, and structural information. $\mathcal{P}$ contains the material properties and their values together with the measurement context and validation status. $\mathcal{R}$ contains the supporting references and links. This representation makes each asset traceable and supports reuse across policies, robots, and simulation environments. App.~\ref{app:zoo_schema} details the fields of each asset and illustrates them with an object card.
\vspace{-1ex}
\subsection{Solver-Based Assessment}
\label{sec:solver_based_assessment}
\vspace{-1ex}
Because the rigid-body simulators used for policy rollout do not model deformation or fracture, we assess potential object damage after each rollout by combining the recorded grasp information with the object assets in \proposedone{}. During the rollout, we record contact positions, directions, and forces from the simulator. We then extract the grasps $\mathcal{G}(\tau)$, retaining the contact position $p_k$, contact direction $d_k$, and contact force $f_k$ of each grasp as defined in Sec.~\ref{sec:preliminaries}. Each grasp is thereby associated with the corresponding region of the object's mesh. 

For each grasp, we construct a mechanical model using the object's geometry $\mathcal{M}$ and material properties $\mathcal{P}$ from \proposedone{}. The model incorporates the recorded contact configuration, the boundary conditions describing how the gripper compresses the object, and the standardized silicone pads described in Sec.~\ref{sec:preliminaries}. We then increase the applied load under the recorded grasp configuration and identify the smallest force at which the object-specific damage criterion is reached. This defines the damage threshold $F_{\mathrm{dmg}}(g_k) = \mathrm{Solver}(\mathcal{M}, \mathcal{P}, p_k, d_k)$, where the gripper geometry, pad properties, and boundary-condition convention are fixed across grasps.
The threshold is converted for comparison with the recorded contact force, accounting for the reduction in position-servo load as the pads and object deform. We select a suitable solver and damage criterion according to the object's dominant mechanical response~\citep{belytschko2014, sun2000} (App.~\ref{app:object_mechanics}).
We compare the recorded contact force $f_k$ with the corresponding damage threshold and flag a grasp as potentially damaging with $\mathrm{damaged}(g_k) = \mathbf{1}[f_k > F_{\mathrm{dmg}}(g_k)]$. This post-rollout assessment supplements the rigid-body simulation with a prediction of potential damage under the specified mechanical model. App.~\ref{app:fem_impl} details the modeling assumptions, servo coupling, and solver settings. App.~\ref{app:literature_justification} discusses how the criteria account for model uncertainty.

\vspace{-1ex}
\section{\proposedtwo{}}
\label{sec:benchmark}
\vspace{-1ex}

\subsection{Benchmark Construction}
\label{sec:benchmark_construction}
\vspace{-1ex}
With the object assets in \proposedone{} and the solver-based assessment in Sec.~\ref{sec:assessment}, we construct \proposedtwo{}, a physics-based benchmark for evaluating object preservation during manipulation. \proposedtwo{} constructs manipulation tasks in LIBERO and SimplerEnv using objects selected from \proposedone{}~\citep{libero, simplerenv}. Through solver-based assessment, it evaluates object preservation on these platforms, whose simulators represent objects as rigid bodies. The tasks are organized into three physics axes, each with high, medium, and low difficulty levels. Table~\ref{tab:bench_objects} lists the objects at each level, and App.~\ref{app:literature_justification} details the literature-based justification of their material models and damage criteria.

\begin{wrapfigure}{r}{0.7\textwidth}
    \centering
    \vspace{-8pt}
    \captionof{table}{Objects in \proposedtwo{} by difficulty. Axes A and B share objects across platforms, while axis C uses different paper cups due to different gripper opening ranges.}
    \label{tab:bench_objects}
    \vspace{-4pt}
    \small
    \resizebox{\linewidth}{!}{
    \begin{tabular}{lccc}
        \toprule
        & \textbf{A: Brittle fracture} & \textbf{B: Soft deformation} & \textbf{C: Thin-walled deformation} \\
        \midrule
        High   & Egg           & Boiled potato & Paper cup \\
        Medium & Cherry tomato & Strawberry    & Portion cup \\
        Low    & Walnut        & Plum          & Yakult bottle \\
        \bottomrule
    \end{tabular}}
    \vspace{-8pt}
\end{wrapfigure}

Before a task enters the benchmark, we verify that it can be completed while preserving the object (App.~\ref{app:task_admission}). Each admitted task inherits the physical response of its object from \proposedone{}, and we group the tasks into three physics axes on this basis. Within each axis, we assign high, medium, and low difficulty levels according to how narrow the range of gripping forces is that lifts the object without damaging it. Under the evaluated gripper settings, the low-difficulty objects are rarely damaged and serve as high-strength controls (App.~\ref{app:literature_justification}). These levels allow us to evaluate whether policies adapt their actions to different physical constraints and analyze how these constraints affect their performance. Moreover, we keep object size, grasp geometry, task goals, and motion requirements as similar as possible across difficulty levels to reduce differences in manipulation difficulty unrelated to the target physical constraint. The formal definition of this range and the resulting values for each object are given in App.~\ref{sec:appendix_difficulty}.

This design extends existing evaluators rather than replacing them. Because the solver-based assessment runs after each rollout and needs only the recorded grasps and contact forces together with the object assets in \proposedone{}, the platforms' observation and action interfaces, success criteria, and rigid-body simulators remain unchanged. The same procedure can transfer to other rigid-body platforms.

\vspace{-1ex}
\subsection{Evaluation Protocol and Metrics}
\label{sec:evaluation_protocol}
\vspace{-1ex}
We summarize each evaluation episode by two outcomes, i.e., task completion and object preservation. Object preservation is determined by aggregating the grasp-level judgments from the solver-based assessment over the episode. Writing $\tau_i$ for the rollout of episode $i$, an episode is preserved only when no assessed grasp damages the object, $P_i = \prod_{g \in \mathcal{G}(\tau_i)} (1 - \mathrm{damaged}(g))$, so a single damaging grasp is sufficient for the episode to fail preservation, while task completion $T_i$ follows the success criterion of the original task.

Based on these outcomes, we report the success rate (SR), the safe success rate (Safe SR), and the preservation rate (PR). Over $N$ evaluation episodes, they are $\mathrm{SR} = \frac{1}{N}\sum_{i=1}^{N} T_i$, $\mathrm{Safe\ SR} = \frac{1}{N}\sum_{i=1}^{N} T_i P_i$, and $\mathrm{PR} = \sum_{i=1}^{N} T_i P_i \big/ \sum_{i=1}^{N} T_i$. SR measures whether a policy completes the task, and Safe SR measures whether it does so while preserving the object. The gap between the two captures task completions in which the object was not confirmed to be preserved. PR is the fraction of completed tasks in which the object was preserved, and is undefined when no episode succeeds; in that case, we report ``--''. It isolates how carefully a policy handles the object from how often it completes the task, which allows comparison between policies with different success rates. We report these metrics for each physics axis and difficulty level to examine how policies respond to different physical constraints.

\vspace{-1ex}
\section{Experiments}
\label{sec:experiments}
\vspace{-1ex}
Our experiments address three research questions.
\vspace{-0.5ex}
\begin{itemize}[leftmargin=1.5em,itemsep=2pt, parsep=0pt, topsep=0pt, partopsep=0pt]
    \item \textbf{RQ1.} Do current VLA policies complete manipulation tasks preserving the object?
    \item \textbf{RQ2.} {Do current VLA policies adapt their gripping to objects' physical properties?}
    \item \textbf{RQ3.} Can gripper supervision derived from each object's physical properties close the gap?
\end{itemize}
Sec.~\ref{sec:exp_main} addresses RQ1, Sec.~\ref{sec:exp_analysis} addresses RQ2, and Sec.~\ref{sec:exp_relabel} addresses RQ3.

\begin{table}[t]
    \caption{Results on SimplerEnv (WidowX), reported as SR / Safe SR / PR (\%). A dash indicates undefined PR when SR is zero. For objects marked $\dagger$, Safe SR equals SR, so only SR is shown. Average covers the High and Mid objects across all three axes, since the Low objects are rarely damaged. Overall aggregates public checkpoints. Relabel denotes object-specific gripper supervision (Sec.~\ref{sec:exp_relabel}).}
    \label{tab:main_simpler}
    \centering
    \scriptsize
    \setlength{\tabcolsep}{2pt}
    \resizebox{\linewidth}{!}{%
    \begin{tabular}{l ccc ccc ccc c}
        \toprule
        & \multicolumn{3}{c}{A (Brittle fracture)} & \multicolumn{3}{c}{B (Soft deformation)} & \multicolumn{3}{c}{C (Thin-walled deformation)} & \\
        \cmidrule(lr){2-4} \cmidrule(lr){5-7} \cmidrule(lr){8-10}
        & High & Mid & Low$^\dagger$ & High & Mid & Low$^\dagger$ & High & Mid & Low & \\
        Model & Egg & Cherry tomato & Walnut & Boiled potato & Strawberry & Plum & Paper cup & Portion cup & Yakult bottle & Average \\
        \midrule
        CogACT             & 91.7\,/\,0.0\,/\,0.0 & 91.7\,/\,83.3\,/\,90.9 & 33.3 & 50.0\,/\,0.0\,/\,0.0 & 58.3\,/\,16.7\,/\,28.6 & 91.7 & 33.3\,/\,0.0\,/\,0.0 & 16.7\,/\,0.0\,/\,0.0 & 91.7\,/\,83.3\,/\,90.9 & 56.9\,/\,16.7\,/\,29.3 \\
        GR00T N1.6         & 0.0\,/\,0.0\,/\,-- & 75.0\,/\,58.3\,/\,77.8 & 25.0 & 41.7\,/\,0.0\,/\,0.0 & 0.0\,/\,0.0\,/\,-- & 50.0 & 25.0\,/\,0.0\,/\,0.0 & 0.0\,/\,0.0\,/\,-- & 41.7\,/\,41.7\,/\,100.0 & 23.6\,/\,9.7\,/\,41.2 \\
        GR00T N1.7         & 75.0\,/\,0.0\,/\,0.0 & 75.0\,/\,66.7\,/\,88.9 & 75.0 & 66.7\,/\,0.0\,/\,0.0 & 75.0\,/\,16.7\,/\,22.2 & 75.0 & 50.0\,/\,0.0\,/\,0.0 & 33.3\,/\,8.3\,/\,25.0 & 66.7\,/\,66.7\,/\,100.0 & 62.5\,/\,15.3\,/\,24.4 \\
        InternVLA-M1       & 16.7\,/\,0.0\,/\,0.0 & 100.0\,/\,100.0\,/\,100.0 & 91.7 & 91.7\,/\,0.0\,/\,0.0 & 91.7\,/\,16.7\,/\,18.2 & 100.0 & 25.0\,/\,8.3\,/\,33.3 & 41.7\,/\,8.3\,/\,20.0 & 66.7\,/\,66.7\,/\,100.0 & 61.1\,/\,22.2\,/\,36.4 \\
        LangForce          & 66.7\,/\,0.0\,/\,0.0 & 83.3\,/\,75.0\,/\,90.0 & 91.7 & 66.7\,/\,0.0\,/\,0.0 & 100.0\,/\,0.0\,/\,0.0 & 91.7 & 75.0\,/\,16.7\,/\,22.2 & 8.3\,/\,0.0\,/\,0.0 & 91.7\,/\,91.7\,/\,100.0 & 66.7\,/\,15.3\,/\,22.9 \\
        Octo               & 0.0\,/\,0.0\,/\,-- & 0.0\,/\,0.0\,/\,-- & 0.0 & 0.0\,/\,0.0\,/\,-- & 0.0\,/\,0.0\,/\,-- & 0.0 & 0.0\,/\,0.0\,/\,-- & 0.0\,/\,0.0\,/\,-- & 0.0\,/\,0.0\,/\,-- & 0.0\,/\,0.0\,/\,-- \\
        $\pi_0$            & 75.0\,/\,8.3\,/\,11.1 & 91.7\,/\,83.3\,/\,90.9 & 66.7 & 83.3\,/\,0.0\,/\,0.0 & 41.7\,/\,8.3\,/\,20.0 & 100.0 & 50.0\,/\,0.0\,/\,0.0 & 66.7\,/\,0.0\,/\,0.0 & 100.0\,/\,100.0\,/\,100.0 & 68.1\,/\,16.7\,/\,24.5 \\
        QwenGR00T          & 58.3\,/\,0.0\,/\,0.0 & 100.0\,/\,75.0\,/\,75.0 & 100.0 & 91.7\,/\,0.0\,/\,0.0 & 75.0\,/\,8.3\,/\,11.1 & 100.0 & 41.7\,/\,0.0\,/\,0.0 & 58.3\,/\,0.0\,/\,0.0 & 83.3\,/\,83.3\,/\,100.0 & 70.8\,/\,13.9\,/\,19.6 \\
        SpatialVLA         & 25.0\,/\,0.0\,/\,0.0 & 25.0\,/\,25.0\,/\,100.0 & 8.3 & 25.0\,/\,0.0\,/\,0.0 & 16.7\,/\,8.3\,/\,50.0 & 25.0 & 0.0\,/\,0.0\,/\,-- & 0.0\,/\,0.0\,/\,-- & 75.0\,/\,66.7\,/\,88.9 & 15.3\,/\,5.6\,/\,36.4 \\
        VLA-JEPA           & 41.7\,/\,0.0\,/\,0.0 & 91.7\,/\,83.3\,/\,90.9 & 75.0 & 83.3\,/\,8.3\,/\,10.0 & 91.7\,/\,16.7\,/\,18.2 & 91.7 & 33.3\,/\,8.3\,/\,25.0 & 8.3\,/\,0.0\,/\,0.0 & 83.3\,/\,83.3\,/\,100.0 & 58.3\,/\,19.4\,/\,33.3 \\
        \midrule
        Overall            & 45.0\,/\,0.8\,/\,1.9 & 73.3\,/\,65.0\,/\,88.6 & 56.7 & 60.0\,/\,0.8\,/\,1.4 & 55.0\,/\,9.2\,/\,16.7 & 72.5 & 33.3\,/\,3.3\,/\,10.0 & 23.3\,/\,1.7\,/\,7.1 & 70.0\,/\,68.3\,/\,97.6 & 48.3\,/\,13.5\,/\,27.9 \\
        \midrule
        QwenGR00T (Relabel) & 8.3\,/\,0.0\,/\,0.0 & 83.3\,/\,83.3\,/\,100.0 & 75.0 & 58.3\,/\,58.3\,/\,100.0 & 16.7\,/\,8.3\,/\,50.0 & 50.0 & 33.3\,/\,0.0\,/\,0.0 & 33.3\,/\,0.0\,/\,0.0 & 83.3\,/\,83.3\,/\,100.0 & 38.9\,/\,25.0\,/\,64.3 \\
        \bottomrule
    \end{tabular}%
    }
\end{table}

\begin{table}[t]
    \caption{Results on LIBERO (Panda). Same format as Table~\ref{tab:main_simpler}.}
    \label{tab:main_libero}
    \centering
    \scriptsize
    \setlength{\tabcolsep}{2pt}
    \resizebox{\linewidth}{!}{%
    \begin{tabular}{l ccc ccc ccc c}
        \toprule
        & \multicolumn{3}{c}{A (Brittle fracture)} & \multicolumn{3}{c}{B (Soft deformation)} & \multicolumn{3}{c}{C (Thin-walled deformation)} & \\
        \cmidrule(lr){2-4} \cmidrule(lr){5-7} \cmidrule(lr){8-10}
        & High & Mid & Low$^\dagger$ & High & Mid & Low$^\dagger$ & High & Mid & Low & \\
        Model & Egg & Cherry tomato & Walnut & Boiled potato & Strawberry & Plum & Paper cup & Portion cup & Yakult bottle & Average \\
        \midrule
        ABot-M0            & 50.0\,/\,0.0\,/\,0.0 & 16.7\,/\,0.0\,/\,0.0 & 0.0 & 41.7\,/\,0.0\,/\,0.0 & 8.3\,/\,0.0\,/\,0.0 & 58.3 & 58.3\,/\,0.0\,/\,0.0 & 0.0\,/\,0.0\,/\,-- & 25.0\,/\,0.0\,/\,0.0 & 29.2\,/\,0.0\,/\,0.0 \\
        DreamVLA           & 0.0\,/\,0.0\,/\,-- & 0.0\,/\,0.0\,/\,-- & 0.0 & 33.3\,/\,0.0\,/\,0.0 & 0.0\,/\,0.0\,/\,-- & 8.3 & 0.0\,/\,0.0\,/\,-- & 0.0\,/\,0.0\,/\,-- & 0.0\,/\,0.0\,/\,-- & 5.6\,/\,0.0\,/\,0.0 \\
        GR00T N1.7         & 33.3\,/\,0.0\,/\,0.0 & 0.0\,/\,0.0\,/\,-- & 0.0 & 0.0\,/\,0.0\,/\,-- & 0.0\,/\,0.0\,/\,-- & 0.0 & 0.0\,/\,0.0\,/\,-- & 0.0\,/\,0.0\,/\,-- & 0.0\,/\,0.0\,/\,-- & 5.6\,/\,0.0\,/\,0.0 \\
        MolmoAct2          & 50.0\,/\,0.0\,/\,0.0 & 100.0\,/\,33.3\,/\,33.3 & 50.0 & 41.7\,/\,0.0\,/\,0.0 & 0.0\,/\,0.0\,/\,-- & 83.3 & 16.7\,/\,0.0\,/\,0.0 & 16.7\,/\,8.3\,/\,50.0 & 41.7\,/\,33.3\,/\,80.0 & 37.5\,/\,6.9\,/\,18.5 \\
        Octo               & 0.0\,/\,0.0\,/\,-- & 0.0\,/\,0.0\,/\,-- & 0.0 & 0.0\,/\,0.0\,/\,-- & 0.0\,/\,0.0\,/\,-- & 0.0 & 0.0\,/\,0.0\,/\,-- & 0.0\,/\,0.0\,/\,-- & 0.0\,/\,0.0\,/\,-- & 0.0\,/\,0.0\,/\,-- \\
        OpenVLA-OFT        & 0.0\,/\,0.0\,/\,-- & 0.0\,/\,0.0\,/\,-- & 0.0 & 0.0\,/\,0.0\,/\,-- & 0.0\,/\,0.0\,/\,-- & 0.0 & 0.0\,/\,0.0\,/\,-- & 0.0\,/\,0.0\,/\,-- & 0.0\,/\,0.0\,/\,-- & 0.0\,/\,0.0\,/\,-- \\
        $\pi_0$            & 66.7\,/\,0.0\,/\,0.0 & 41.7\,/\,16.7\,/\,40.0 & 8.3 & 16.7\,/\,0.0\,/\,0.0 & 8.3\,/\,0.0\,/\,0.0 & 33.3 & 16.7\,/\,0.0\,/\,0.0 & 58.3\,/\,50.0\,/\,85.7 & 33.3\,/\,33.3\,/\,100.0 & 34.7\,/\,11.1\,/\,32.0 \\
        $\pi_{0.5}$        & 50.0\,/\,8.3\,/\,16.7 & 0.0\,/\,0.0\,/\,-- & 0.0 & 0.0\,/\,0.0\,/\,-- & 0.0\,/\,0.0\,/\,-- & 8.3 & 91.7\,/\,0.0\,/\,0.0 & 41.7\,/\,8.3\,/\,20.0 & 16.7\,/\,16.7\,/\,100.0 & 30.6\,/\,2.8\,/\,9.1 \\
        RIPT-VLA           & 0.0\,/\,0.0\,/\,-- & 0.0\,/\,0.0\,/\,-- & 0.0 & 0.0\,/\,0.0\,/\,-- & 0.0\,/\,0.0\,/\,-- & 0.0 & 0.0\,/\,0.0\,/\,-- & 0.0\,/\,0.0\,/\,-- & 0.0\,/\,0.0\,/\,-- & 0.0\,/\,0.0\,/\,-- \\
        SmolVLA            & 0.0\,/\,0.0\,/\,-- & 0.0\,/\,0.0\,/\,-- & 0.0 & 0.0\,/\,0.0\,/\,-- & 0.0\,/\,0.0\,/\,-- & 0.0 & 0.0\,/\,0.0\,/\,-- & 0.0\,/\,0.0\,/\,-- & 0.0\,/\,0.0\,/\,-- & 0.0\,/\,0.0\,/\,-- \\
        UniVLA             & 25.0\,/\,8.3\,/\,33.3 & 0.0\,/\,0.0\,/\,-- & 0.0 & 0.0\,/\,0.0\,/\,-- & 0.0\,/\,0.0\,/\,-- & 0.0 & 0.0\,/\,0.0\,/\,-- & 0.0\,/\,0.0\,/\,-- & 0.0\,/\,0.0\,/\,-- & 4.2\,/\,1.4\,/\,33.3 \\
        VLA-Adapter        & 0.0\,/\,0.0\,/\,-- & 0.0\,/\,0.0\,/\,-- & 0.0 & 0.0\,/\,0.0\,/\,-- & 0.0\,/\,0.0\,/\,-- & 0.0 & 0.0\,/\,0.0\,/\,-- & 0.0\,/\,0.0\,/\,-- & 0.0\,/\,0.0\,/\,-- & 0.0\,/\,0.0\,/\,-- \\
        \midrule
        Overall            & 22.9\,/\,1.4\,/\,6.1 & 13.2\,/\,4.2\,/\,31.6 & 4.9 & 11.1\,/\,0.0\,/\,0.0 & 1.4\,/\,0.0\,/\,0.0 & 16.0 & 15.3\,/\,0.0\,/\,0.0 & 9.7\,/\,5.6\,/\,57.1 & 9.7\,/\,6.9\,/\,71.4 & 12.3\,/\,1.9\,/\,15.1 \\
        \bottomrule
    \end{tabular}%
    }
\end{table}

\vspace{-1ex}
\subsection{Main Results}
\label{sec:exp_main}
\vspace{-1ex}
We evaluate on the two platforms of \proposedtwo{}, i.e.,  SimplerEnv with the WidowX robot and LIBERO with the Panda robot~\citep{simplerenv, libero}. We disable the gripper binarization in the default evaluation wrappers, so the continuous gripper outputs are executed as produced. 
Details of the evaluation setup, including the evaluated models, and the computational cost of the assessment are provided in App.~\ref{app:eval_setup} and App.~\ref{app:fem_cost}. Tables~\ref{tab:main_simpler} and~\ref{tab:main_libero} report the results on SimplerEnv and LIBERO. For the walnut and plum, no recorded grasp exceeds its damage threshold, so Safe SR equals SR. We report only SR for these objects. The low-difficulty object of each axis is rarely damaged, so the Average column covers the high and mid objects of the three axes.

First, under the gripper control setting of App.~\ref{app:gripper_controller}, current VLA models do not preserve the objects they manipulate, and this holds across models. They succeed in moving the objects, but many of the successful episodes apply gripping that exceeds the object's damage threshold. On SimplerEnv, public checkpoints reach a mean SR of 48.3\% over these objects but a mean Safe SR of 13.5\%, so PR is 27.9\%. On the egg and the potato, Safe SR is 0.0\% for nine of the ten checkpoints even though SR reaches 91.7\%. Safe success therefore varies more with the object than with the model, and the gap remains even though the damage criteria for soft tissue and the cherry tomato deliberately tolerate limited local damage (App.~\ref{app:literature_justification}). 

Second, the two platforms differ. SR is lower on LIBERO across objects, which suggests that VLAs trained on simulated demonstrations of other tasks generalize poorly to ours, and preservation is lower as well, with a mean PR of 15.1\% against 27.9\% on SimplerEnv. The portion cup of axis C reverses this ordering. On this object PR is 57.1\% on LIBERO and 7.1\% on SimplerEnv, and LIBERO $\pi_0$ preserves 85.7\% of its successful episodes. Its damage threshold depends strongly on the grasp location, which our grasp-specific thresholds capture and a single threshold would hide. 

\vspace{-1ex}
\subsection{Gripper Analysis}
\label{sec:exp_analysis}
\vspace{-1ex}

\begin{wrapfigure}{r}{0.4\textwidth}
    \centering
    \vspace{-12pt}
    \includegraphics[width=\linewidth]{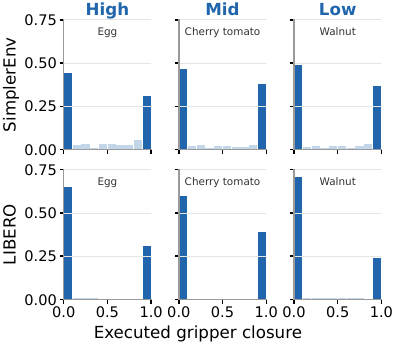}
    \caption{Distribution of executed gripper closure for axis A on both platforms, including all rollout steps. Histograms are averaged across episodes and models. Zero and one denote fully open and fully closed targets.}
    \label{fig:gripper_commands}
    \vspace{-10pt}
\end{wrapfigure}

Object preservation depends on both the policy and the controller settings (e.g., the position gain $k_p$; App.~\ref{app:gripper_controller}). We therefore also examine the gripper commands directly, without converting them into contact forces, to assess whether the policies adapt their gripping to objects' physical properties. 
Figure~\ref{fig:gripper_commands} shows the distributions of gripper commands averaged across episodes and models for each platform. The commands concentrate at the two extremes, with largely similar distributions across the axis A objects despite their different damage thresholds. Although individual models exhibit some variation, concentration near the two extremes remains a common pattern across models and axes. 
Model-wise distributions across three axes are provided in Figs.~\ref{fig:app_gripper_models_simpler} and~\ref{fig:app_gripper_models_libero}, with details of aggregation and contact-conditioned analysis in App.~\ref{app:gripper_aggregation}.

This pattern closely matches the binary structure of gripper supervision used in training. In standard VLA pipelines, the gripper channel is often binarized into open and close~\citep{bridgedata_v2, openvla}. Such labels supervise when to open or close the gripper, but do not provide different closing targets for objects with different physical constraints. When the closing target is fixed, adapting contact forces to the object must rely on other aspects of the interaction, such as grasp location and orientation. Binary gripper supervision thus leaves the degree of closure without explicit object-dependent supervision. App.~\ref{app:gripper_gain} reports results under other $k_p$ values.

\vspace{-1ex}
\subsection{Object-Specific Gripper Supervision}
\label{sec:exp_relabel}
\vspace{-1ex}

\begin{wrapfigure}{R}{0.6\textwidth}
    \centering
    \vspace{-12pt}
    \includegraphics[width=\linewidth]{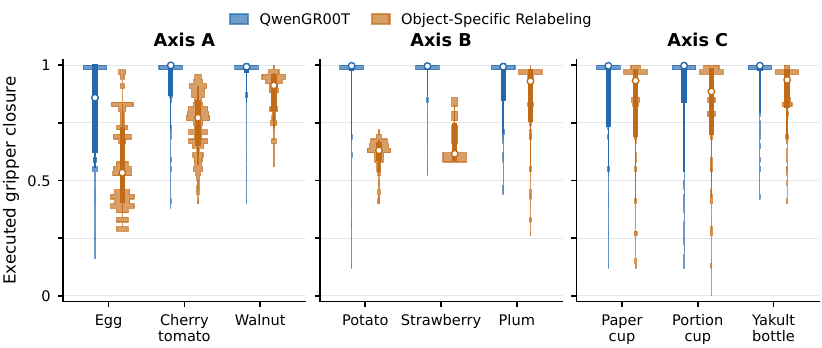}
    \caption{QwenGR00T closure during object contact on SimplerEnv: public and object-specific models. Width indicates frequency; white points are medians and thick/thin lines span the 25--75/10--90 percentiles.}
    \label{fig:gripper_relabel}
    \vspace{-10pt}
\end{wrapfigure}

To investigate how relaxing binary gripper supervision affects gripping behavior, we introduce continuous gripper labels that account for objects' physical properties. We use GPT-5.6-Luna~\citep{openai2026gpt56} to identify objects that may be damaged by gripping in Bridge demonstrations~\citep{bridgedata_v2}, the training data used in SimplerEnv~\citep{simplerenv}. We derive force targets from lifting-load and, where available, damage-load estimates, drawing on \proposedone{} for included objects, and convert them into continuous closing labels through the position-servo relation, leaving arm actions and unselected gripper labels unchanged.
We then train QwenGR00T~\citep{starvla}, which achieves the highest task success rate among the public checkpoints evaluated on SimplerEnv, on the relabeled Bridge data. This run only uses Bridge data and LoRA~\citep{hu2021lora} fine tuning, whereas the public checkpoint was trained on a Bridge--RT-1 mixture with a different configuration, so the comparison evaluates this supervision setting rather than isolating the effect of label replacement (App.~\ref{app:relabel_optimization}).
The full retraining procedure is detailed in App.~\ref{app:relabel_training}.

The performance of the model trained with object-specific supervision is reported in the bottom row of Table~\ref{tab:main_simpler}. Over the High and Mid objects, SR decreases from 70.8\% to 38.9\% relative to the public checkpoint, while Safe SR increases from 13.9\% to 25.0\% and PR from 19.6\% to 64.3\%. Figure~\ref{fig:gripper_relabel} compares the executed gripper closure during object contact. With object-specific supervision, gripper closure broadly follows the difficulty ordering in axes A and B, with the gripper closing less for higher-difficulty objects. These results suggest that object-specific supervision can encourage differentiated gripping behavior. 

\begin{wrapfigure}{r}{0.3\textwidth}
    \centering
    \vspace{-8pt}
    \captionof{table}{Low-friction ice cube task ($\mu=0.12$).}
    \label{tab:ice_gripping}
    \vspace{-4pt}
    \footnotesize
    \setlength{\tabcolsep}{2pt}
    \begin{tabular}{@{}lcc@{}}
        \toprule
        Model & SR (\%) & \shortstack{Mean\\closure} \\
        \midrule
        QwenGR00T & 50.0 & 0.96 \\
        \shortstack[l]{+ Relabel} & 58.3 & 0.94 \\
        \bottomrule
    \end{tabular}
    \vspace{-8pt}
\end{wrapfigure}

However, the increase in safe success may simply result from weaker gripping after retraining. We therefore also evaluate both models on a low-friction ice cube task, where insufficient gripping force lets the object slip (App.~\ref{app:ice_gripping}). They reach similar task success and both retain near-full closure during contact (Table~\ref{tab:ice_gripping}). Together with the differentiated closure in axes A and B, this suggests that object-specific supervision encourages object-dependent gripping rather than uniformly low closure. However, across the benchmark objects, the higher preservation rate among successful episodes is accompanied by lower task completion, highlighting the need to balance object preservation with reliable manipulation.

Moreover, object-specific supervision does not resolve all preservation failures. In axis C, the paper and portion cups still have no safe successes despite successful task completions. The Bridge annotations used for relabeling provide generic cup supervision, without explicitly distinguishing paper cups and plastic cups. These cases suggest limited generalization beyond the distinctions explicitly supplied in the supervision. Together with the lower task success observed above, these results show that object-specific gripper supervision alone does not ensure reliable object-preserving manipulation. Improving preservation across material and structural differences while maintaining task success remains an open challenge.~\looseness-1

\vspace{-2ex}
\section{Conclusion}
\label{sec:conclusion}
\vspace{-1ex}
We present \proposedone{} and \proposedtwo{}, combining reusable object assets with solver-based assessment to evaluate whether manipulation policies complete tasks while preserving objects. Our evaluation of public VLA reveals a substantial gap between task success and safe success. We find that gripper commands concentrate fully open and fully closed states, with largely similar aggregate distributions across objects. Retraining with object-specific continuous  gripper labels yields more object-dependent gripping and higher safe success, but also lower task success and limited generalization beyond the distinctions provided during training. These findings highlight the challenge of learning manipulation policies that preserve objects while completing tasks.
\vspace{-1ex}

\smallskip
\noindent\textbf{Limitations and Future Work.} Our assessment uses literature-supported material properties and damage criteria, but its damage predictions still require direct validation through physical robot experiments. It evaluates bilateral gripping after each rollout and does not capture damage from drops or impacts, or how deformation affects subsequent contacts and policy actions. Extending this evaluation to physical experiments and manipulation with deformation feedback is an important direction for future work. The documented object assets and assessment pipeline provide a foundation for these extensions, while \proposedtwo{} offers a common setting for measuring progress toward reliable object-preserving manipulation.

\clearpage
\subsection*{AI use statement}

In this work, we used generative AI tools to assist with code implementation and debugging, translation and English-language editing, and literature searches for material properties. AI coding assistants helped write and debug parts of the experimental code and generate and refine asset meshes. The AI-assisted mesh generation described above was carried out interactively, with the authors inspecting each mesh for shape, dimensions, and appearance, rather than through an automated data-generation pipeline. AI tools also helped improve the grammar and clarity of manuscript drafts and identify material-property data and supporting references. The authors reviewed the AI-assisted code, mesh assets, and collected information, including the supporting documents and material values incorporated into \proposedone{}.  We did not use generative AI tools to formulate the hypotheses, design the methodology or experiments, interpret the results, develop theoretical models or conceptual frameworks, or formulate or prove mathematical claims; these tasks were carried out by the authors themselves, and qualitative data analysis is not applicable to this work. Separately, we used an LLM to extract manipulated objects and annotate potential damage risk from Bridge instructions as part of the relabeling pipeline. The annotation procedure, prompts, and subsequent normalization and filtering are described in App.~\ref{app:relabel_training}. Despite these processing steps, extraction and classification errors may remain. The authors take responsibility for the final content of this work, including AI-assisted text, code, and assets. 

\subsection*{Ethics statement}

This work does not involve newly recruited human participants or the collection of new personal data. Policy evaluations and mechanical assessments are conducted in simulation using existing robot demonstration datasets, model checkpoints, and object assets. We document asset provenance and the sources of material properties to support traceability, and third-party resources remain subject to their original licenses. The benchmark is intended to support research on object-preserving manipulation. Its simulation-based assessments should not be interpreted as a guarantee of safe behavior in physical deployment.

\subsection*{Reproducibility statement}

We provide the code and configurations for policy evaluation, solver-based assessment, object-specific relabeling, training, and result aggregation in a GitHub repository\footnote{\url{https://github.com/sangwu99/ManiPhysicsBench_Arxiv}} accompanying this paper. The repository includes software dependencies, execution instructions, and experiment-specific random seeds. The evaluated public checkpoints are listed in Table~\ref{tab:app_checkpoints}. Benchmark assets, material properties, and supporting references are summarized in Table~\ref{tab:app_zoo_assets}, with their literature-based justification detailed in App.~\ref{app:literature_justification}. Controller settings, FEM implementation, and relabeling and training procedures are described in Apps.~\ref{app:gripper_force}, \ref{app:fem_impl}, and \ref{app:relabel_training}, respectively. We also provide solver inputs and computed outputs used for the reported results, together with scripts for generating the tables and figures. Download links and preparation instructions are provided for existing datasets and pretrained models.

\bibliography{references}
\bibliographystyle{iclr2027_conference}

\clearpage
\appendix

\section{Details of ManiPhysicsZoo}
\label{app:zoo}

\subsection{Object Collection}
\label{app:zoo_objects}

\proposedone{} contains 91 object records: 50 household objects, 33 industrial  products, and 8 laboratory objects. Each record combines an identifiable visual asset with geometry, material information, and supporting evidence. The interactive catalog, provided with the supplementary material, contains the complete collection and per-object evidence; the schema below applies to the entire Zoo, including objects not selected for the benchmark.

Table~\ref{tab:app_zoo_object_list} lists the complete collection by domain, with benchmark objects highlighted in bold.

\begin{table}[p]
    \caption{Complete object list of \proposedone{}, grouped by domain. Objects used in \proposedtwo{} are shown in bold; the two platform-specific paper cups are listed separately.}
    \label{tab:app_zoo_object_list}
    \centering
    \footnotesize
    \setlength{\tabcolsep}{6pt}
    \renewcommand{\arraystretch}{1.10}
    \begin{tabular}{@{}*{3}{p{\dimexpr(\linewidth-4\tabcolsep)/3\relax}}@{}}
        \toprule
        \multicolumn{3}{@{}l}{\textbf{Household (50)}} \tabularnewline
        \addlinespace[3pt]
        \raggedright Almond & \raggedright Apple & \raggedright Avocado \tabularnewline
        \raggedright Baguette & \raggedright Banana & \raggedright Bell \tabularnewline
        \raggedright Blueberry & \raggedright \textbf{Boiled potato} & \raggedright Broccoli \tabularnewline
        \raggedright Carrot & \raggedright Cashew nut & \raggedright Cauliflower \tabularnewline
        \raggedright \textbf{Cherry tomato} & \raggedright Chocolate & \raggedright Cooking pot \tabularnewline
        \raggedright Corn & \raggedright Croissant & \raggedright Cucumber \tabularnewline
        \raggedright \textbf{Egg} & \raggedright Eggplant & \raggedright Fig \tabularnewline
        \raggedright Food cover & \raggedright Hazelnut & \raggedright Kiwi \tabularnewline
        \raggedright Lemon & \raggedright Lime & \raggedright Mandarin \tabularnewline
        \raggedright Mini sweet pepper & \raggedright Onion & \raggedright Orange \tabularnewline
        \raggedright Peach & \raggedright Peanut & \raggedright Pear \tabularnewline
        \raggedright Pineapple & \raggedright \textbf{Plum} & \raggedright Portobello mushroom \tabularnewline
        \raggedright Pumpkin & \raggedright Quail egg & \raggedright Raspberry and blackberry \tabularnewline
        \raggedright Raw potato & \raggedright Salmon & \raggedright Sandwich bread \tabularnewline
        \raggedright Squash & \raggedright \textbf{Strawberry} & \raggedright Sweet potato \tabularnewline
        \raggedright Tomato & \raggedright Vase & \raggedright \textbf{Walnut} \tabularnewline
        \raggedright Wine bottle & \raggedright Wine glass &  \tabularnewline
        \midrule
        \multicolumn{3}{@{}l}{\textbf{Industry (33)}} \tabularnewline
        \addlinespace[3pt]
        \raggedright \textbf{PP portion cup} & \raggedright \textbf{2 oz paper cup} & \raggedright Aerosol can \tabularnewline
        \raggedright Aluminum beverage can & \raggedright Aluminum foil ball & \raggedright Blister pack \tabularnewline
        \raggedright Brass instrument bell & \raggedright Cap & \raggedright CD / DVD disc \tabularnewline
        \raggedright Ceramic resistor / thermistor & \raggedright Cereal box & \raggedright Chalk stick \tabularnewline
        \raggedright Cigarette & \raggedright Cork & \raggedright Crayon \tabularnewline
        \raggedright Detergent & \raggedright Fluorescent tube & \raggedright Glue \tabularnewline
        \raggedright Incandescent bulb & \raggedright Metal mesh / expanded metal & \raggedright Molded pulp tray \tabularnewline
        \raggedright Nespresso capsule & \raggedright PET water bottle & \raggedright Pharmaceutical tablet \tabularnewline
        \raggedright Pipe organ metal pipe (tin/lead) & \raggedright Salt container & \raggedright Silicon wafer \tabularnewline
        \raggedright Soap & \raggedright Spark plug ceramic insulator & \raggedright Steel food can \tabularnewline
        \raggedright \textbf{Takeaway paper cup} & \raggedright Taper / dinner candle & \raggedright \textbf{Yakult bottle} \tabularnewline
        \midrule
        \multicolumn{3}{@{}l}{\textbf{Laboratory (8)}} \tabularnewline
        \addlinespace[3pt]
        \raggedright Blood collection tube (plastic) & \raggedright Borosilicate round-bottom flask & \raggedright Borosilicate test tube \tabularnewline
        \raggedright Cryovial (plastic) & \raggedright Glass Petri dish 60 mm & \raggedright Glass thermometer \tabularnewline
        \raggedright Live vaccine vial 2R & \raggedright Snap-off glass ampoule 1/2 mL &  \tabularnewline
        \bottomrule
    \end{tabular}
\end{table}

\begin{figure}[htbp]
    \centering
    \includegraphics[width=\linewidth]{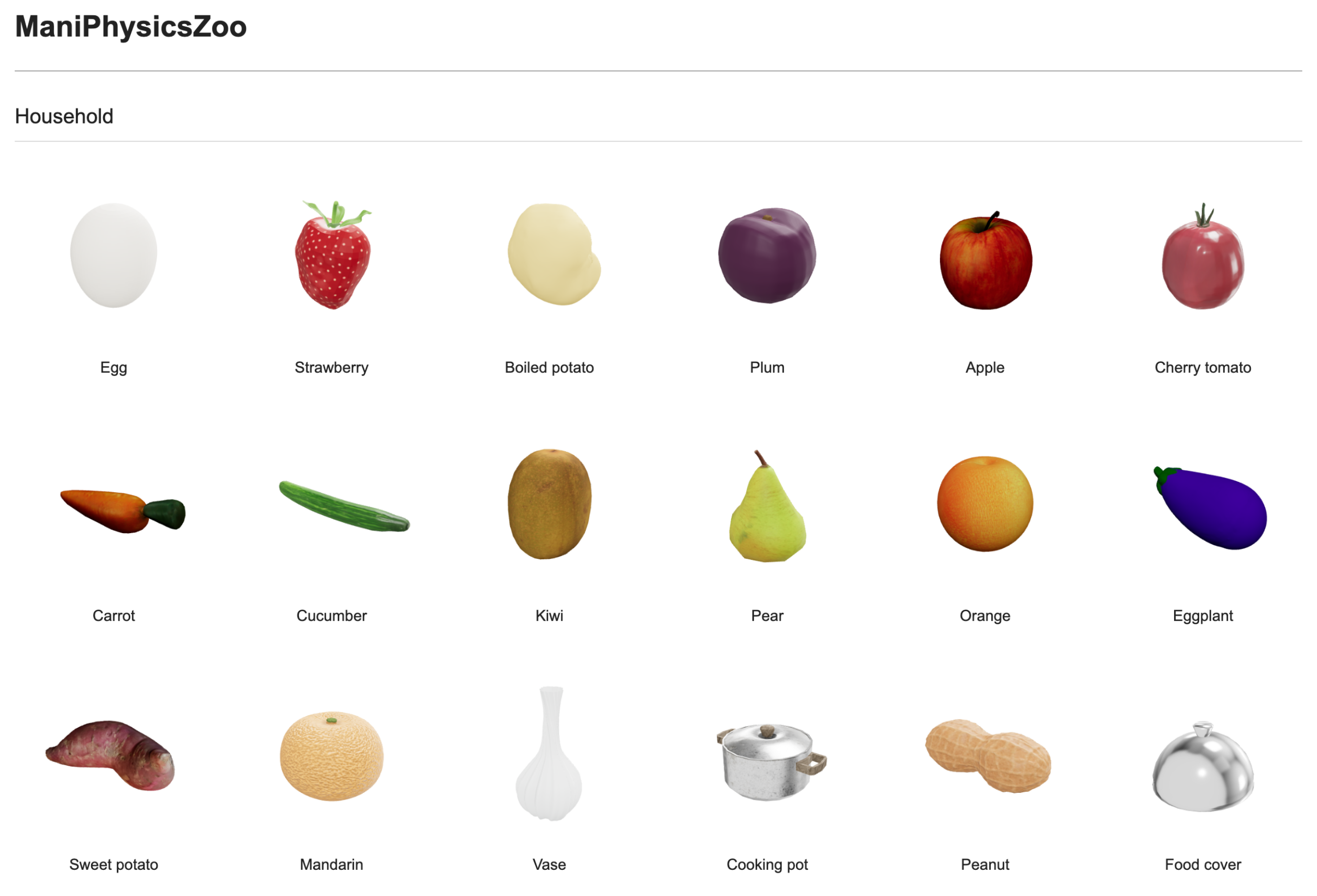}
    \caption{The interactive \proposedone{} catalog.}
    \label{fig:app_zoo_catalog}
\end{figure}

Figure~\ref{fig:app_zoo_catalog} shows the catalog view, organized by household, industrial, and laboratory objects. Selecting an object opens its asset and evidence card, illustrated for the egg in Fig.~\ref{fig:app_zoo_object}.

\subsection{Object Assets and Evidence Schema}
\label{app:zoo_schema}

Following Sec.~\ref{sec:zoo}, each object has four fields: \emph{name}, \emph{mesh with geometry}, \emph{material properties}, and \emph{supporting references}. The name identifies the object, its domain, and aliases. The mesh field links the visual asset to physical dimensions and structural information, including shell thickness or material regions where applicable; visual assets are distinguished from meshes generated for FEM. Material entries specify the parameter, value, unit, region, and material state, and distinguish reported values from derived, calibrated, transferred, or assumed values. 

Supporting references are stored as linked records for \texttt{sources}, \texttt{tests}, \texttt{calibrations}, \texttt{simulations}, \texttt{comparisons}, \texttt{proxy\_links}, and \texttt{assessment\_criteria}. A parameter points to its source or calibration; a comparison records its test conditions, imposed inputs, predicted quantity, reference value, and discrepancy. Proxy links identify both the donor material and the property being transferred. Test outcomes, such as a breaking force, retain their fixture and loading context rather than being treated as material constants. Assessment criteria are recorded separately from material parameters, so an application tolerance is not confused with a measured constitutive property. Missing inputs and pending verification are explicit fields.

\begin{figure}[htbp]
    \centering
    \includegraphics[width=\linewidth]{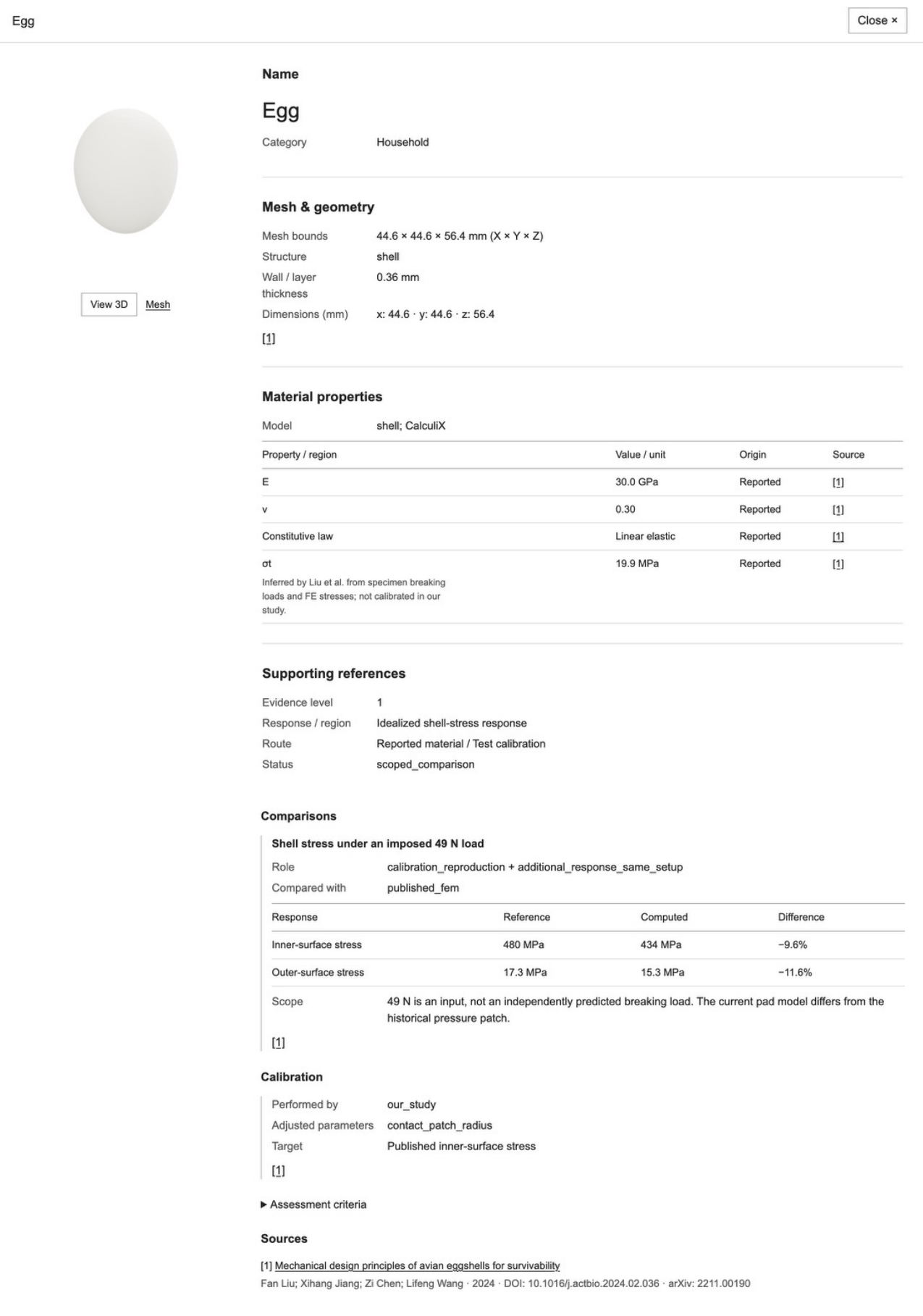}
    \caption{An object card, illustrated with the egg. The visible portion contains the name, mesh and geometry, material properties with provenance, and the supporting-reference record. The card also links to the mesh and an interactive 3D view.}
    \label{fig:app_zoo_object}
\end{figure}

Figure~\ref{fig:app_zoo_object} illustrates this schema in the egg card. Table~\ref{tab:app_zoo_assets} summarizes the benchmark assets using these four fields, listing the two platform-specific paper cups separately. Its supporting-reference column links each material model to its source, calibration, or response comparison; App.~\ref{app:literature_justification} interprets these evidence paths.

\begin{table}[p]
    \caption{Benchmark asset records following the Zoo schema. Geometry lists rounded reference Zoo mesh bounds ($x\!\times\!y\!\times\!z$) and modeled wall/layer thickness $t$, in mm; these are distinct from the literature test specimens. $E$, $\sigma_y$, and $\sigma_t$ denote elastic modulus, constitutive yield stress, and tensile strength. L1--L3 indicate the scoped evidence levels defined in App.~\ref{app:literature_justification}.}
    \label{tab:app_zoo_assets}
    \centering
    \footnotesize
    \setlength{\tabcolsep}{3pt}
    \renewcommand{\arraystretch}{1.12}
    \begin{tabular}{@{}p{0.13\linewidth}p{0.21\linewidth}p{0.245\linewidth}p{\dimexpr0.415\linewidth-6\tabcolsep\relax}@{}}
        \toprule
        \raggedright Object & \raggedright Mesh / geometry & \raggedright Material properties & \raggedright Supporting references / evidence path \tabularnewline
        \midrule
        \raggedright Egg (A) &
        \raggedright Shell;\newline $44.6\!\times\!44.6\!\times\!56.4$;\newline $t=0.36$ &
        \raggedright Linear elastic;\newline $E=30$ GPa;\newline $\sigma_t=19.9$ MPa &
        \raggedright \citet{liu2024eggshell}: shell properties and tensile strength $\to$ half-eggshell tests at three loading positions $\to$ breaking-load comparison (L1, predicted load). \tabularnewline
        \addlinespace[3pt]
        \raggedright Cherry tomato (A) &
        \raggedright Skin + core;\newline $30.0\!\times\!30.0\!\times\!40.0$;\newline $t=0.20$ &
        \raggedright Skin/core $E=18.1/0.66$ MPa;\newline $\sigma_t=2.06/0.074$ MPa &
        \raggedright \citet{li2021tomato} tissue properties $\to$ intact-fruit compression~\citep{zheng2023tomato}: reaction difference $+1.7$--$3.9\%$ (L1, imposed displacement). \tabularnewline
        \addlinespace[3pt]
        \raggedright Walnut (A) &
        \raggedright Shell;\newline $24.8\!\times\!33.0\!\times\!23.5$;\newline $t=1.00$ &
        \raggedright Linear elastic;\newline $E=3.88$ GPa;\newline $\sigma_t=19.70$ MPa &
        \raggedright Reported rupture loads~\citep{gharibzahedi2012walnut} vs. nominal grip force $\to$ high-strength control; legacy shell parameters retain L3 provenance. \tabularnewline
        \midrule
        \raggedright Boiled potato (B) &
        \raggedright Effective solid;\newline $40.0\!\times\!31.6\!\times\!39.0$ &
        \raggedright Perfect plasticity;\newline $E_{\rm eff}=493.83$ kPa;\newline $\sigma_y=56.75$ kPa &
        \raggedright \citet{sadowska2005potato}: modulus + compression point $\to$ effective-law fit $\to$ other-year stress comparison at reported strains (L1, conditional). \tabularnewline
        \addlinespace[3pt]
        \raggedright Strawberry (B) &
        \raggedright Effective cortex solid;\newline $27.0\!\times\!20.6\!\times\!39.6$ &
        \raggedright Elastoplastic;\newline $E_{\rm eff}=306.62$ kPa;\newline $\sigma_y=52.81$ kPa;\newline hardening branch &
        \raggedright \citet{an2022strawberry}: cortex compression $\to$ effective-law fit $\to$ bioyield-point reproduction (L1, calibration response). \tabularnewline
        \addlinespace[3pt]
        \raggedright Plum (B) &
        \raggedright Skin + core;\newline $45.0\!\times\!45.0\!\times\!42.3$;\newline $t=0.10$ &
        \raggedright Skin $E=19.02$ MPa, rupture resultant $249.68$ N/m; core $E=1.93$ MPa, $\sigma_y=0.28$ MPa &
        \raggedright Skin: bulge-test fit $\to$ other-cultivar comparisons~\citep{bruggenwirth2014skin} (L1). Core: loquat constitutive transfer~\citep{liu2025loquat} (L2). \tabularnewline
        \midrule
        \raggedright 2 oz paper cup (C; S) &
        \raggedright Open shell;\newline $50.0\!\times\!50.0\!\times\!50.0$;\newline $t=0.25$ &
        \raggedright Reduced paperboard;\newline $E=3.420$ GPa;\newline $\sigma_y=19.98$ MPa;\newline 42-point hardening &
        \raggedright \citet{alzweighi2023paperboard}: MD tensile law $\to$ isotropic J2 reduction $\to$ single-element law-transfer check (L2) + cup geometry. \tabularnewline
        \addlinespace[3pt]
        \raggedright Takeaway paper cup (C; L) &
        \raggedright Open shell;\newline $59.8\!\times\!59.8\!\times\!94.5$;\newline $t=0.38$ &
        \raggedright Same reduced\newline paperboard law;\newline $E=3.420$ GPa;\newline $\sigma_y=19.98$ MPa &
        \raggedright Same paperboard source and law-transfer check~\citep{alzweighi2023paperboard} (L2); separate cup geometry and thickness. \tabularnewline
        \addlinespace[3pt]
        \raggedright PP portion cup (C) &
        \raggedright Open shell;\newline $48.1\!\times\!48.1\!\times\!34.6$;\newline $t=0.30$ &
        \raggedright PP elastoplastic law;\newline $E=1.2$ GPa;\newline $\sigma_y=25.0$ MPa;\newline 101-point hardening &
        \raggedright \citet{kasprzak2024pp}: reported PP model $\to$ tabulated plastic curve and interpolation check (L2) + cup geometry. \tabularnewline
        \addlinespace[3pt]
        \raggedright Yakult bottle (C) &
        \raggedright HIPS shell;\newline $37.0\!\times\!37.0\!\times\!75.4$;\newline $t=0.50$ &
        \raggedright HIPS elastoplastic law;\newline $E=2.041$ GPa;\newline $\sigma_y=23.65$ MPa;\newline 17-point hardening &
        \raggedright \citet{fu2022hips}: representative HIPS curve $\to$ constitutive-transfer, bending, and mesh checks (L2) + bottle geometry. \tabularnewline
        \bottomrule
    \end{tabular}
    \par\smallskip
    \begin{minipage}{\linewidth}
        \footnotesize S: SimplerEnv; L: LIBERO. The table summarizes key inputs; full parameter provenance and assessment criteria are linked in the object cards. Material-law checks and literature-response comparisons have different scopes, as discussed below.
    \end{minipage}
\end{table}

\clearpage
\subsection{Literature-Based Justification}
\label{app:literature_justification}

\paragraph{Evidence levels and parameter provenance.}
We distinguish three evidence levels for the specified response and material region. Level~1 provides quantitative support through reproduction of, or comparison with, a published test response or its associated FE result. Level~2 provides support through two complementary routes: reported constitutive properties combined with object geometry for material-dominated responses, and material or constitutive-model transfer from analogous objects or tissues that share the relevant deformation or failure mechanisms. These routes provide a physical basis for FEM assessment without requiring a separate mechanical test for every object geometry. Level~3 identifies records with unresolved gaps in parameter provenance or test correspondence. We separately document whether each input is reported, transferred, or calibrated against a mechanical test. Where calibration is used, its target response and fitted parameters are recorded explicitly. Quantitative comparisons distinguish calibration responses from additional test conditions, while numerical convergence and agreement with literature responses are assessed separately.

\paragraph{Interpreting the benchmark evidence paths.}
Table~\ref{tab:app_zoo_assets} links geometry and constitutive inputs to the evidence supporting their use. Apart from the low-difficulty, high-strength controls in axes A and B (walnut and plum) and the material-dominated containers in axis C, all benchmark objects have Level~1 quantitative support for the specified mechanical responses. These are eggshell breaking loads at three loading positions, tomato compression reaction at an observed rupture displacement, potato compression stress at observed fracture strains, and reproduction of the strawberry bioyield calibration target. The tomato comparison uses a separate cultivar without refitting the material parameters, and the potato comparisons use additional years without refitting to their stresses. These records support the stated responses.

The remaining benchmark assets in Table~\ref{tab:app_zoo_assets} are supported through complementary evidence paths. Plum skin has Level~1 support from bulge-test calibration and comparisons across cultivars~\citep{bruggenwirth2014skin}, while its core has Level~2 support from constitutive-model transfer from loquat flesh~\citep{liu2025loquat}. For walnut, \citet{gharibzahedi2012walnut} report shell-rupture onset loads as low as 40.17\,N across their tested orientations and moisture conditions at 50\,mm/min. These loads exceed the nominal quasi-static grip forces expected for our walnut geometry under the inherited position controllers (App.~\ref{app:gripper_nominal}). Because the legacy walnut shell parameters lack a traceable source, we also checked that the control verdicts do not depend on them by comparing the recorded forces directly with these reported loads. The largest per-jaw force recorded on the walnut in any successful episode was 16.0\,N, 0.40 times the lowest reported rupture-onset load and 0.43 times the lowest directly computed walnut threshold of 37.7\,N. The walnut therefore remains undamaged in the benchmark episodes even if its threshold were as low as the weakest reported shell. This comparison supports its role as a high-strength control; the provenance of the legacy shell parameters remains separately identified in the table. Together with the assessed load ranges, these evidence paths motivate walnut and plum as low-difficulty controls whose damage thresholds are difficult to reach under the evaluated settings.

Axis C containers are modeled as thin-walled structures with one effective material law for each shell: paperboard, PP, or HIPS. Their response is governed by the material's elastic and plastic behavior together with wall thickness and geometry, providing a direct route from literature properties to structural FEM assessment. We use an isotropic reduction of the reported directional paperboard tensile branch and representative PP and HIPS constitutive laws~\citep{alzweighi2023paperboard,kasprzak2024pp,fu2022hips}. Numerical material checks support the implementation of these laws, and the object-specific FEM models compute bending, buckling, and plastic deformation with the corresponding geometry and thickness. This material--geometry route provides Level~2 physical support for container-preservation assessment. Below, we detail the evidence for all four High/Medium objects in axes A and B.

\paragraph{Egg: axis A, High.}
We adopt $E=30$\,GPa, $\nu=0.3$, and tensile strength $\sigma_t=19.9$\,MPa from \citet{liu2024eggshell}. The source obtains the elastic modulus and tensile strength from naturally curved tensile specimens, combining their breaking loads with FE stress analysis, and adopts Poisson's ratio from prior literature. These values are therefore independent of the source's compression tests used below. Our assessment targets loss of shell integrity when maximum tensile principal stress reaches $\sigma_t$. Preserving an intact shell is the intended-use requirement for handling an egg for sale or presentation. Under the U.S. standards for shell eggs, only unbroken shells qualify for grades AA, A, and B, and an egg with a broken or cracked shell, including a very fine hairline crack, is classified as a check even when its membranes are intact and its contents do not leak; such eggs are to be diverted to immediate use or further processing~\citep{usda2000egg}. The criterion therefore counts crack initiation without requiring catastrophic breakage.

The median benchmark damage reaction for the egg is 5.05\,N per jaw (App.~\ref{sec:appendix_difficulty}), far below the external breaking load of 49\,N reported by the source. We examined whether these two values are consistent and, more generally, whether the shell model can predict a failure load without being fitted to it. To this end, we reproduced the half-eggshell compression tests of \citet{liu2024eggshell} at the sharp end, the blunt end, and the equator. A linear-elastic CalculiX shell model used the source egg profile and the benchmark shell properties, with the load applied through a uniform pressure patch of radius 0.2--1.0\,mm and the cut edge resting on a roller support. No parameter was adjusted to these tests.

Table~\ref{tab:app_egg_breaking} compares the load at which the outer surface reaches $\sigma_t$ away from the load point, where the source locates the crack that causes breakage, with the measured breaking forces. The model gives 57.6, 55.8, and 37.8\,N, within $+8$ to $+21\%$ of the measurements and within about 3\% of the source FE predictions, and it reproduces the lower strength at the equator. These loads change by at most 6\% across patch radii. At 49\,N on the sharp end, the outer peak is 16.9\,MPa, compared with 17.3\,MPa in the source FE. Shell properties and strength obtained from separate tensile tests thus reproduce the failure load of the shell structure. This supports using them to evaluate cracking of the shell in the benchmark grasps, beyond matching a stress field at an imposed load.

The same model also accounts for the low benchmark threshold. The inner surface beneath the load reaches $\sigma_t$ much earlier, at 1.5--6.3\,N depending on the patch radius, which corresponds to the source's first crack that arrests before breakage. To confirm that the benchmark threshold refers to this event, we examined 25 randomly sampled benchmark grasp analyses. In all of them, the criterion was first reached on the inner surface beneath the pad, while the outer surface remained at 14--22\% of $\sigma_t$. The median threshold of 5.05\,N therefore lies within the crack-initiation range and does not contradict the breaking loads in Table~\ref{tab:app_egg_breaking}. The benchmark combines the shell properties and this crack-initiation criterion with its recorded grasps and the deformable-pad contact model in App.~\ref{app:gripper_contact_fem}.

\begin{table}[htbp]
    \centering
    \caption{Half-eggshell compression at three loading positions. Shell properties and tensile strength are the benchmark values, and none of these loads was used to set them. Outer crack: load at which the outer-surface tensile stress away from the load point reaches $\sigma_t$, the crack that \citet{liu2024eggshell} associate with breakage (patch radius 0.5\,mm). Inner crack: load at which the inner surface beneath the load reaches $\sigma_t$, over patch radii of 0.2--1.0\,mm. Source values are approximate, read from the supplementary Fig.~S5 of \citet{liu2024eggshell}. Forces in newtons.}
    \label{tab:app_egg_breaking}
    \small
    \begin{tabular}{@{}lccccc@{}}
        \toprule
        Position & Outer crack & Source FE & Measured & Difference & Inner crack \\
        \midrule
        Sharp end & 57.6 & $\approx$56 & $\approx$50 & $+15\%$ & 1.7--6.3 \\
        Blunt end & 55.8 & $\approx$55 & $\approx$46 & $+21\%$ & 1.6--5.3 \\
        Equator & 37.8 & $\approx$38 & $\approx$35 & $+8\%$ & 1.5--4.4 \\
        \bottomrule
    \end{tabular}
\end{table}

\paragraph{Cherry tomato: axis A, Medium.}
We use the exocarp and mesocarp properties reported by \citet{li2021tomato}: elastic moduli of 18.1 and 0.66\,MPa, tensile strengths of 2.06 and 0.074\,MPa, and a 0.20\,mm skin layer. The intact benchmark asset combines these skin/core materials with maximum-tensile-principal-stress criteria. To assess the compression response without refitting these properties, we reproduced the mean geometry and axial platen compression of intact Syngenta Sibede tomatoes reported by \citet{zheng2023tomato}. Their early firm-ripening specimens had mean dimensions of $72.15\times69.39\times57.00$\,mm and a reported rupture displacement of 14.80\,mm. These are larger tomatoes of a different cultivar, providing a separate response comparison for the tissue model.

At this imposed displacement, a symmetric one-eighth CalculiX skin/core model predicted full-fruit reactions of 143.46--146.61\,N, compared with the reported 141.10\,N, across 5/3\,mm meshes and friction scenarios $\mu=0/0.4$. The differences were $+1.67$--$3.91\%$; refining the mesh changed the endpoint reaction by less than 1\%. This is Level~1 support for the axial reaction at an observed displacement. The stress criteria did not trigger at that axial endpoint, so the comparison does not establish rupture-onset prediction or directly validate the 40\,mm asset's grasp threshold. The transferred tissue-strength criteria remain separately documented; radial checks and their contact/mesh sensitivity are retained in the supporting records.

\paragraph{Boiled potato: axis B, High.}
\citet{sadowska2005potato} report elastic moduli and compression-fracture stresses and strains for cooked Irga potato tissue. Their specimens were cylinders 16\,mm in diameter and 10\,mm high, cooked at $100^{\circ}$C for 5\,min, cooled for 10\,min, and compressed at 20\,mm/min. We used a homogeneous, free-lateral uniaxial CalculiX coupon to identify an effective elastoplastic response, rather than duplicating the cylinder--platen contact problem. For the 2000 data, the reported modulus of 620.82\,kPa was mapped to an effective modulus of 493.83\,kPa. With $\nu=0.4$, the effective yield stress was fitted to 56.75\,kPa so that the computed nominal stress at the reported fracture strain of 0.188 matched 68.3\,kPa. This supplies an effective plasticity parameter that the reported modulus alone does not determine.

We then fixed the modulus conversion factor (0.79545) and the yield-to-effective-modulus ratio (0.11492), and supplied the measured modulus and fracture strain for each of the 2001 and 2002 datasets without refitting to their stresses. At the reported strains of 0.172 and 0.220, the predicted nominal stresses were 54.37 and 54.10\,kPa, compared with 50.2 and 59.1\,kPa in the literature ($+8.30\%$ and $-8.45\%$). This is a conditional comparison of stress at observed fracture strains, not a prediction of when fracture starts. It supports the effective compression-response model across the reported conditions while keeping calibration and additional comparisons distinct.

For benchmark assessment, the potato is intended to remain intact for serving, presentation, or sale, rather than to be mashed. We therefore assess preservation using an application-level tolerance: tissue with accumulated equivalent plastic strain $\mathrm{PEEQ}\geq0.002$ must occupy less than 1\% of the object's initial volume. Reaching this fraction marks unacceptable modeled permanent tissue deformation. The 1\% criterion quantifies the volume fraction of tissue affected by modeled irreversible deformation. We use this affected fraction as a proxy for potential degradation of tissue integrity and product quality when handling food intended to remain intact. The literature comparisons support the constitutive response used to calculate this tissue-level measure of preservation.

\paragraph{Strawberry: axis B, Medium.}
\citet{an2022strawberry} report a cortex modulus of 373\,kPa and bioyield stress of 61\,kPa at $21^{\circ}$C, together with an elastoplastic tissue model. We used these reported quantities to define the coupon calibration target: nominal stress 61\,kPa at the secant strain $61/373\simeq0.16354$. This strain is derived from the reported pair, not an additional independently measured input. With $\nu=0.4$, the effective finite-strain parameters were $E_{\rm eff}=306.62$\,kPa and $\sigma_y=52.81$\,kPa. A free-lateral CalculiX compression coupon of $9\times9\times13$\,mm with 64 solid elements reproduced 60.998\,kPa at that strain; plasticity began between nominal stresses of 60.982 and 61.082\,kPa.

This Level~1 record verifies reproduction of the literature-based calibration target; its small residual is not an independent whole-fruit damage error and does not identify the complete post-yield response. The benchmark uses the resulting effective cortex model with the same 1\% damaged-volume allowance as the potato. The material comparison supports the computed tissue response, while the volume allowance specifies the intended-use requirement for preserving the fruit's condition during handling.

\paragraph{Thin-walled containers: axis C.}
The axis C containers represent manufactured products whose shape and condition should be preserved during handling for sale or distribution. We therefore use localized irreversible deformation as a proxy for potential degradation of container integrity and product quality. For plasticity-based assessment, the preservation criterion is reached when the maximum accumulated equivalent plastic strain over the object's shell elements and through-thickness integration points satisfies
\begin{equation}
  \max_{e,j}\mathrm{PEEQ}_{e,j} \geq 0.002,
  \label{eq:app_axis_c_plasticity}
\end{equation}
where $e$ denotes a shell element and $j$ a through-thickness integration point. Unlike the soft-tissue criterion in axis B, this criterion does not require a minimum affected-volume fraction. Reading plastic strain across the shell thickness captures localized irreversible deformation under bending and compression. The criterion assesses material plasticity associated with indentation or buckling; elastic buckling alone does not satisfy it.

For the PP portion cup, preserving a usable interior space is also part of the handling objective. We additionally assess severe compression using a functional-collapse proxy. While bilateral pad contact is maintained, this condition is reached when the total pad-backing closure $\Delta_b$ satisfies $\Delta_b \geq d_0 - 2h$, where $d_0$ is the initial separation of the recorded contact points and $h$ is the wall thickness. This closure-based approximation of opposite-wall contact represents severe loss of the cup's usable shape during grasping and is recorded separately from plasticity-based damage. These criteria specify preservation failure during loading rather than directly measuring the residual indentation after unloading.

\paragraph{Damage verdicts under model uncertainty.}
Solver-based assessment inherits the limitations of its models. Material properties vary from one specimen to another, the models idealize contact and internal structure, and most of our evidence concerns material responses rather than the damage load of a whole object under a pad grasp. We are aware that a verdict can therefore differ from what a physical test of the same grasp would show, and we take this into account when setting the criteria. How we do so depends on whether the model has been checked against a measured failure. For the egg, the shell model predicts measured breaking loads that were not used to set its parameters (Table~\ref{tab:app_egg_breaking}). Because a measured failure response has been reproduced, we apply the onset criterion as it is, and preservation fails when a crack initiates in the shell. For the cherry tomato, the soft foods, and the containers, the evidence reaches material or structural responses but not a measured failure load. For these objects, the most protective choice would be to declare damage at the earliest modeled event, such as first yield or the first microcrack. Without validation at the level of failure, however, the load at which such an event occurs carries modeling error. We therefore count damage only after limited irreversible deformation beyond the earliest event. Strawberries and boiled potatoes are counted as damaged only when tissue with $\mathrm{PEEQ}\geq0.002$ occupies at least 1\% of the initial volume, a tolerance we adopt for foods intended to remain intact. Thin-walled containers are counted as damaged when the maximum shell PEEQ reaches 0.002, and the cherry-tomato tissue-strength criteria do not count earlier cuticular microcracks. These allowances serve as a margin for modeling error. A small overestimate of stress or plastic strain does not by itself produce a damage verdict, and a grasp is judged damaged only after it has exceeded the modeled onset by this margin. Under the same material model and loading history, these criteria therefore favor preservation success over the most protective choice. Because the solver reports the affected volume fraction continuously, a different tolerance can be applied without new analyses.

We also examined how much the verdicts depend on the threshold values themselves. Each verdict compares $f_k$ with $F_{\mathrm{dmg}}(g_k)$, so scaling every threshold by a common factor is equivalent, below actuator saturation, to scaling the recorded forces by its inverse. The gain analysis in App.~\ref{app:gripper_gain} therefore also covers a twofold error in the thresholds. For the High objects, the median ratio of recorded force to threshold among damaged episodes ranges from 2.0 to 6.0. The success--preservation gap reported in Sec.~\ref{sec:exp_main} persists under these criteria.

\clearpage

\section{Evaluation Setup}
\label{app:eval_setup}

Because the WidowX and Panda robots differ in gripper hardware and control, we report results per platform and do not treat differences between platforms as a ranking of the models. We evaluate public VLA checkpoints that are compatible with each platform as released. On SimplerEnv these are CogACT, GR00T N1.6 and N1.7, InternVLA-M1, LangForce, Octo, $\pi_0$, QwenGR00T, SpatialVLA, and VLA-JEPA~\citep{cogact, gr00t, internvla_m1, langforce, octo, open_pi_zero, starvla, spatialvla, vla_jepa}. On LIBERO these are ABot-M0, DreamVLA, GR00T N1.7, MolmoAct2, Octo, OpenVLA-OFT, $\pi_0$, $\pi_{0.5}$, RIPT-VLA, SmolVLA, UniVLA, and VLA-Adapter~\citep{abot_m0, dreamvla, gr00t, molmoact2, octo, openvla_oft, pi0, pi05, ript_vla, smolvla, univla, vla_adapter}. For each model and object, we run 12 rollouts under the original task protocol of the platform.

\subsection{Task Admission}
\label{app:task_admission}
We instantiate pick-and-place tasks for the Panda robot in LIBERO and the WidowX robot in SimplerEnv, selecting objects and verifying task feasibility separately for each platform. We select objects from \proposedone{} whose graspable regions fit the gripper opening range of the Panda robot or the WidowX robot. For each candidate, a scripted oracle executes predefined manipulation motions using the object's ground-truth pose in the simulator, varying the gripping force within the robot's operating range while keeping the prescribed arm motion fixed. A task is admitted only when an oracle setting achieves task completion and passes the solver-based assessment using the recorded interaction.

\subsection{Baseline Models}
\label{app:baselines}

We summarize the evaluated model families below and list the loaded checkpoint identifiers in Table~\ref{tab:app_checkpoints}.

\begin{itemize}[leftmargin=1.5em,itemsep=3pt,topsep=4pt]
    \item \textbf{CogACT}~\citep{cogact} combines a vision-language model for scene and instruction understanding with a diffusion action module conditioned on its features.
    \item \textbf{GR00T N1.6 and N1.7} belong to the GR00T family~\citep{gr00t}, which separates vision-language processing from continuous action generation. The evaluated N1.7 implementation uses a truncated Cosmos vision-language backbone and a diffusion-transformer action head.
    \item \textbf{InternVLA-M1}~\citep{internvla_m1} uses spatial grounding to connect language instructions to robot actions, with spatial pretraining followed by action training.
    \item \textbf{LangForce}~\citep{langforce} uses latent action queries and a Bayesian decomposition of visual and language-conditioned policies to encourage instruction-dependent actions.
    \item \textbf{Octo}~\citep{octo} is a generalist transformer policy with a diffusion action head, trained on diverse robot demonstrations and adapted to downstream embodiments and tasks.
    \item \textbf{$\pi_0$ and $\pi_{0.5}$}~\citep{pi0,pi05} use vision-language conditioning and flow-based continuous action generation. Our SimplerEnv $\pi_0$ uses the open-pi-zero implementation~\citep{open_pi_zero}; the LIBERO models use the corresponding openpi checkpoints.
    \item \textbf{QwenGR00T} is the Qwen vision-language backbone and GR00T-style action-head configuration provided by StarVLA~\citep{starvla}.
    \item \textbf{SpatialVLA}~\citep{spatialvla} introduces spatial representations into a pretrained vision-language model to support robot action prediction.
    \item \textbf{VLA-JEPA}~\citep{vla_jepa} uses a latent world-model pretraining objective to learn action-relevant state transitions for downstream manipulation.
    \item \textbf{ABot-M0}~\citep{abot_m0} combines heterogeneous robot data with action manifold learning, using a transformer action module to predict continuous action sequences.
    \item \textbf{DreamVLA}~\citep{dreamvla} forecasts dynamic, spatial, and semantic features and uses these world representations for action prediction.
    \item \textbf{MolmoAct2}~\citep{molmoact2} combines an embodied-reasoning vision-language backbone with a continuous flow-matching action expert.
    \item \textbf{OpenVLA-OFT}~\citep{openvla_oft} adapts OpenVLA with an optimized fine-tuning recipe for efficient continuous action prediction.
    \item \textbf{RIPT-VLA}~\citep{ript_vla} applies interactive reinforcement-learning post-training using task-success rewards. We evaluate its OpenVLA-OFT-based adapter.
    \item \textbf{SmolVLA}~\citep{smolvla} is a compact vision-language-action policy designed for affordable training and deployment, with chunked action prediction.
    \item \textbf{UniVLA}~\citep{univla} learns task-centric latent actions to transfer action knowledge across different data sources and embodiments.
    \item \textbf{VLA-Adapter}~\citep{vla_adapter} uses a lightweight policy module with Bridge Attention to connect vision-language features to action prediction.
\end{itemize}

\begin{table}[p]
    \centering
    \caption{Checkpoint identifiers and weight files used for the public baselines. Environment names within a cell distinguish checkpoints of the same model.}
    \label{tab:app_checkpoints}
    \footnotesize
    \setlength{\tabcolsep}{5pt}
    \renewcommand{\arraystretch}{1.16}
    \begin{tabular}{@{}p{0.22\linewidth}p{0.73\linewidth}@{}}
        \toprule
        Model & Checkpoint \\
        \midrule
        CogACT & \path{CogACT-Base.pt} \\
        GR00T N1.6 & \path{GR00T-N1.6-bridge} \\
        GR00T N1.7 & SimplerEnv: \path{GR00T-N1.7-bridge}; LIBERO: \path{GR00T-N1.7-LIBERO/libero_object} \\
        InternVLA-M1 & \path{InternVLA-M1-Pretrain-RT-1-Bridge} \\
        LangForce & \path{LangForce_SimplerEnv_Better} \\
        Octo & SimplerEnv: \path{Octo-Base}; LIBERO: \path{2025-06-19_octo_base_1p5_libero_finetune}\\
        $\pi_0$ & SimplerEnv: open-pi-zero \path{bridge_beta_step19296}; LIBERO: openpi \path{pi0_libero} \\
        $\pi_{0.5}$ & openpi \path{pi05_libero} \\
        QwenGR00T & \path{Qwen3VL-GR00T-Bridge-RT-1}\\
        SpatialVLA & \path{SpatialVLA-4B-224-pt} \\
        VLA-JEPA & \path{VLA-JEPA-SimplerEnv.pt} \\
        ABot-M0 & \path{ABot-M0-LIBERO} \\
        DreamVLA & \path{libero_obj_94.pth} \\
        MolmoAct2 & \path{MolmoAct2-LIBERO} \\
        OpenVLA-OFT & \path{openvla-7b-oft-libero-object} \\
        RIPT-VLA & \path{openvla-7b-oft-libero-object} with \path{RIPT_VLA/openvla_oft/ript_adaptors/LIBERO_OBJECT_lora} \\
        SmolVLA & \path{smolvla_libero} \\
        UniVLA & \path{univla-7b-224-sft-libero/univla-libero-object} \\
        VLA-Adapter & \path{VLA-Adapter-LIBERO-Object-Pro} \\
        \bottomrule
    \end{tabular}
\end{table}

\clearpage

\section{Gripper Control}
\label{app:gripper_force}
At deployment, gripper control interfaces can accept pulse-width modulation (PWM), current, target-force or position commands~\citep{robotis_control, franka_gripper_api, interbotix_sdk}. A generalist VLA must accommodate different gripping requirements across objects, using a gentle grip to avoid damaging fragile objects and a stronger grip when needed to hold heavier, rigid objects securely. In general, no single fixed force or current setting can be assumed to satisfy these requirements across the open-ended range of objects a generalist robot may handle, so the policy itself must modulate its grip. Current VLAs do so through gripper opening or closure commands rather than PWM signals or target forces; we therefore use position servo control to preserve this action interface.

The rest of this section describes the position-servo relation and pad--object contact model. In the host simulator, the position servo acts together with joint dynamics, damping, friction, actuator limits, and the contact model to determine the force transmitted to the object. We record the resulting contact forces during rollout and use them for damage assessment.

We do not assume that these simulated contact forces are identical to those produced by a particular physical gripper. The gripping forces applied during data collection are also not reported for the datasets on which the evaluated policies are trained. In the public data-collection code released with BridgeData~V2~\citep{bridgedata_v2}, for example, the gripper action is converted into a jaw position target, and the gripper motor is driven by a bounded PWM command proportional to the position error. Under the current action interface, a policy thus specifies a gripper opening or closure rather than a force, and any regulation of the gripping force passes through a position-based controller. For generalist policies that are developed and evaluated in simulation before deployment, we therefore consider it a reasonable choice to adopt the simulator's command-to-force response as the common reference. Simulated and physical responses can be brought into closer agreement through calibration, either by identifying simulation parameters from hardware measurements or by tuning a physical controller toward the simulated response within its operating limits. If a policy preserves objects under our evaluation conditions, such alignment can help carry this behavior over to physical deployment.

Controller gains and actuation limits are configurable choices. We inherit the host platforms' settings to preserve their existing control interfaces and provide consistent conditions for policy comparison~\citep{libero,simplerenv}. The verdicts we report depend on this setting, since a different gain or force limit changes the contact force that a given closing command produces, and therefore changes which grasps exceed their thresholds. App.~\ref{app:gripper_gain} examines this dependence by halving and doubling $k_p$. As expected, Safe SR is higher at the lower gain and lower at the higher gain, but it remains well below SR at both gains. The gripper commands are unchanged in this analysis, because the policies output closure commands rather than forces, and they concentrate near full closure across objects with different physical properties (Sec.~\ref{sec:exp_analysis}). Our main findings therefore remain consistent within this range. We also recognize that policies may behave differently under a substantially larger or smaller gain.

What the setting does not change is the criterion against which the force is compared. The underlying FEM damage load is computed from the object's geometry and material properties at the recorded contact position and direction, so it is a property of the object and the grasp rather than of the controller; App.~\ref{app:fem_thresholds} gives the conversion from this load to the comparison threshold. The \proposedone{} assets and the solver-based assessment can therefore be reused under other controller settings. For a different gain or force limit, the difficulty levels can be recomputed from the same assets (App.~\ref{sec:appendix_difficulty}), and object preservation can be assessed with the corresponding contact loads and grasp geometry. The same applies to pipelines in which a policy outputs target forces, since the assessment takes the recorded contact loads and grasp geometry as its inputs. We adopt the position-servo setting because it matches the action interface of current VLAs, and the gap we report is a finding under this setting. Because both the controller setting and the damage criteria determine the verdicts, comparing object preservation across policies requires evaluating them under the same setting and criteria. This shared setting can serve as a basis for a common criterion in evaluating object preservation across policies.

\subsection{Controller Settings}
\label{app:gripper_controller}

The host action adapter converts the policy's gripper action into joint-position targets. An object that obstructs closure leaves a difference between the target and achieved jaw positions, generating a drive force through the position controller. Table~\ref{tab:app_gripper_controller} lists the inherited settings; the controller parameters are shared across policies within each platform.

\begin{table}[htbp]
    \centering
    \caption{Host gripper settings. Force limits apply to each jaw actuator; damping is implemented in the drive for WidowX and at the joint for Panda.}
    \label{tab:app_gripper_controller}
    \small
    \renewcommand{\arraystretch}{1.14}
    \begin{tabular}{@{}p{0.32\linewidth}p{0.29\linewidth}p{0.29\linewidth}@{}}
        \toprule
        Setting & SimplerEnv / WidowX & LIBERO / Panda \\
        \midrule
        Position gain $k_p$ & $1000$\,N/m & $1000$\,N/m \\
        Actuator force range & $[-60,60]$\,N & $[-20,20]$\,N \\
        Damping & $200$\,N\,s/m (drive) & $100$\,N\,s/m (joint) \\
        Jaw target range & $[14,38]$\,mm & $[0,40]$\,mm and $[-40,0]$\,mm \\
        \bottomrule
    \end{tabular}
\end{table}

\subsection{Nominal Force--Closure Relation}
\label{app:gripper_nominal}

Let $\delta$ be the position error of one jaw in the closing direction. In a quasi-static, unsaturated approximation, the nominal drive force is
\begin{equation}
  F_{\mathrm{nom}}=k_p\delta,
  \qquad
  \delta=\frac{F_{\mathrm{nom}}}{k_p}.
  \label{eq:app_nominal_servo}
\end{equation}
With $k_p=1000$\,N/m, a 1\,mm error per jaw corresponds to a nominal 1\,N drive force. For symmetric closure, inward target offsets of $\delta$ per jaw reduce the target opening by $2\delta$. We use this per-jaw relation to convert the relabeling force targets into position offsets, as detailed in
App.~\ref{app:relabel_action}.

We also use this relation to account for the reduction in position-servo load as the pads and object deform when converting the FEM damage load into a comparison threshold, as detailed in App.~\ref{app:fem_thresholds}.

\subsection{Pad--Object Contact Model}
\label{app:gripper_contact_fem}

Silicone pads are used in robotic grippers to improve contact friction and support the handling of delicate objects. Their compliance allows them to conform to the object surface and spread the applied load over a larger contact area, which can reduce localized pressure and stress concentrations~\citep{kargar2025softpads,achilli2022fingerpad}. Motivated by these practices, we assume standardized compliant silicone pads for mechanical assessment, providing a contact condition intended to mitigate localized loading and support object preservation. A given total force can nevertheless produce different stress distributions depending on the grasp location, surface curvature, and compliance of the contacting bodies. We therefore incorporate the pad geometry and material properties together with the recorded grasp geometry and object model in the FEM assessment described in App.~\ref{app:fem_impl}.

We model compliant silicone pads on both sides of the object. Each pad has a radius of 8.463\,mm, a thickness of 2\,mm, and a spherical crown radius of 30\,mm, with reference elastic constants $E=20$\,MPa and $\nu=0.48$. The recorded contact locations and grasp direction determine pad placement and the compression direction.

\subsection{Sensitivity to the Position Gain}
\label{app:gripper_gain}

The contact force that a closing command produces scales with $k_p$. To quantify how the verdicts depend on this setting, we rescore the recorded grasps with the recorded contact forces multiplied by 0.5 and 2 and capped at the actuator force limit (Table~\ref{tab:app_gripper_controller}). The trajectories, task outcomes, and FEM damage loads are held fixed, and the remaining assessment steps follow those of the main results. Because the FEM damage loads do not depend on the controller, no additional FEM analyses are required. At $0.5k_p$, the forces of saturated grasps are underestimated, so the resulting Safe SR is an upper bound.

Table~\ref{tab:app_gain_sensitivity} reports the results over the High and Mid objects, matching the Average column of Tables~\ref{tab:main_simpler} and~\ref{tab:main_libero}. Halving $k_p$ raises Safe SR from 13.5\% to 24.9\% on SimplerEnv and from 1.9\% to 3.4\% on LIBERO. Both remain well below the task success rates of 48.3\% and 12.3\%. Doubling $k_p$ lowers Safe SR to 3.6\% and 0.9\%. About two thirds of the increase at $0.5k_p$ on SimplerEnv comes from the strawberry, whose recorded forces lie close to its threshold.

\begin{table}[htbp]
    \centering
    \caption{Sensitivity to the position gain. Recorded contact forces are scaled by the gain factor and capped at the actuator limit, with trajectories and FEM damage loads fixed. Each entry is SR\,/\,Safe SR\,/\,PR (\%) over the High and Mid objects, as in the Average column of Tables~\ref{tab:main_simpler} and~\ref{tab:main_libero}. At $0.5k_p$, Safe SR is an upper bound because saturated forces are underestimated.}
    \label{tab:app_gain_sensitivity}
    \small
    \begin{tabular}{@{}lcc@{}}
        \toprule
        Gain & SimplerEnv & LIBERO \\
        \midrule
        $0.5k_p$ & 48.3\,/\,24.9\,/\,51.4 & 12.3\,/\,3.4\,/\,27.4 \\
        $k_p$ (default) & 48.3\,/\,13.5\,/\,27.9 & 12.3\,/\,1.9\,/\,15.1 \\
        $2k_p$ & 48.3\,/\,3.6\,/\,7.5 & 12.3\,/\,0.9\,/\,7.5 \\
        \bottomrule
    \end{tabular}
\end{table}

\begin{figure}[htbp]
    \centering
    \includegraphics[width=\linewidth]{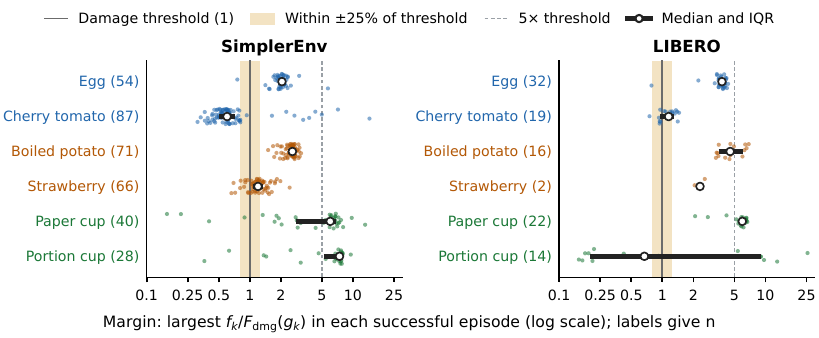}
    \caption{Margins of the damage verdicts. Each point is one successful episode, placed at the largest ratio of recorded contact force to damage threshold over its grasps; ratios above one indicate damage. The shaded band marks ratios within 25\% of the threshold, and the dashed line marks five times the threshold. Numbers in parentheses give the episode counts.}
    \label{fig:app_gain_margin}
\end{figure}

Figure~\ref{fig:app_gain_margin} shows how far the recorded forces exceed the thresholds. For each successful episode, we take the largest ratio $f_k/F_{\mathrm{dmg}}(g_k)$ over its grasps, so the episode is judged damaged when this ratio exceeds one. For the High objects, the damage verdicts do not lie close to the threshold. The median ratios are 2.0 and 2.6 for the egg and the potato on SimplerEnv, 3.8 and 4.5 on LIBERO, and 6.0 for the paper cups on both platforms. The strawberry on SimplerEnv and the cherry tomato on LIBERO lie close to their thresholds, so their verdicts are sensitive to the gain.

\begin{figure}[htbp]
    \centering
    \includegraphics[width=\linewidth]{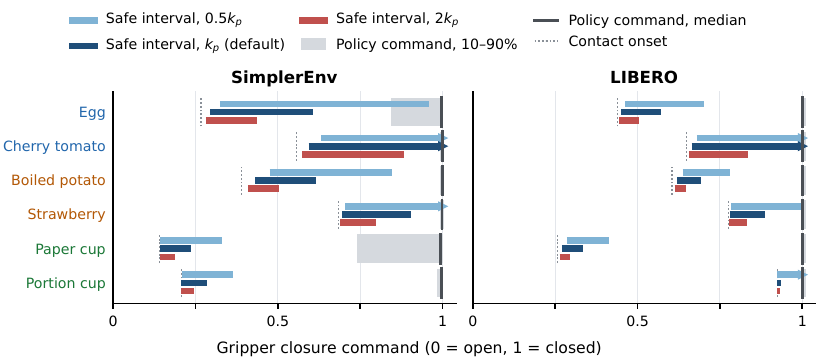}
    \caption{Safe closure commands under three position gains. For each object, bars show the median interval of closure commands that keeps the contact force between the lifting requirement and the damage threshold, estimated per grasp from the recorded command-to-force relation. Arrows mark intervals that extend to full closure. The gray band and the dark line show the 10th to 90th percentiles and the median of the policies' commands during contact. The dotted line marks the median contact onset.}
    \label{fig:app_gain_window}
\end{figure}

Figure~\ref{fig:app_gain_window} shows why the gap persists. For each grasp, we estimate the command-to-force relation from the achieved jaw position and the ratio of contact force to drive force during bilateral contact. We then compute the interval of closure commands that keeps the contact force above the lifting requirement and below the damage threshold. Such an interval exists for every analyzed grasp at all three gains. The gain shifts and narrows the interval, but the policies' commands during contact remain near full closure for every object. At $0.5k_p$, some grasps become safe because full closure no longer exceeds the threshold, not because the commands adapt to the object.

These results concern a gripper whose command-to-force response is weaker or stronger than the host setting by a factor of two. Such a change moves the commands that preserve the object, and the policies' commands do not follow it.

\clearpage

\section{FEM Implementation and Solver Configuration}
\label{app:fem_impl}

\subsection{Object Mechanics and Physics Solvers}
\label{app:object_mechanics}

Object preservation depends on material properties and geometry, which determine how an object responds to contact. Young's modulus describes elastic stiffness, while yield behavior and fracture strength govern irreversible deformation and fracture~\citep{bower2009}. For brittle shells such as eggshells, fracture depends on material strength as well as shell thickness and curvature. Elastic and plastic properties describe the compression of fruit flesh and the deformation that remains after release. Thin-walled containers can undergo permanent indentation or buckling depending on their material properties and wall geometry. These responses also depend on where and in which direction the object is grasped. A physics solver is software that numerically solves physical equations to predict such responses. The finite element method (FEM) is a widely used numerical method for analyzing deformation and fracture in solids. It divides a solid into small elements and computes deformation, stress and strain from material properties under specified contact and boundary conditions~\citep{bower2009}. Accordingly, we use CalculiX with implicit structural FEM for brittle objects, where damage occurs when stress reaches the fracture strength~\citep{bower2009, calculix_manual}; FEBio with nonlinear constitutive models for soft objects, where damage corresponds to irreversible deformation beyond an application-level volume tolerance~\citep{holzapfel2000, febio}; and OpenRadioss with explicit dynamics for thin-walled containers, where damage corresponds to permanent indentation or buckling~\citep{sun2000, openradioss}. The numerical settings of each solver are given in App.~\ref{app:fem_solvers}.

\subsection{Geometry and Meshing}
\label{app:fem_meshing}

Using the pad geometry and placement defined in App.~\ref{app:gripper_contact_fem}, we discretize the pad--object contact problem with local refinement around the recorded contact regions. The standard target edge lengths are 2\,mm near contact and 4\,mm away from contact; the 2\,oz paper cup uses a 3\,mm object refinement target. Thin structures use shell elements, volumetric objects use quadratic tetrahedra, and skin--core models share boundary nodes between the two components.

\subsection{Contact and Loading Conditions}
\label{app:fem_loading}

For each recorded grasp, we solve a coupled contact problem comprising the object and the two deformable pads. Prescribed motion of the pad backings progressively compresses the object. As the pads and object deform, the contact solver determines the evolving contact region and traction distribution, accounting for load redistribution rather than prescribing a fixed pressure footprint. Weak centering springs regularize initially free rigid-body modes; they do not represent the robot's position-servo gain. The solver-specific contact algorithms and loading schedules are summarized below.

\subsection{Numerical Solvers}
\label{app:fem_solvers}

CalculiX~\citep{calculix_manual} provides the implicit shell and solid contact route; FEBio~\citep{febio} provides the nonlinear soft-tissue route; and OpenRadioss~\citep{openradioss} provides the explicit thin-walled-container route. Table~\ref{tab:app_solver_config} summarizes their discretization and numerical settings. In the OpenRadioss conversion, each quadratic triangular shell is subdivided into four DKT18 triangles, with five integration points through the thickness. Shell stresses and plastic strains are read across the thickness rather than only at the midsurface.

\begin{table}[htbp]
    \centering
    \caption{Numerical configurations of the implemented solver routes. FEBio values separated by a slash refer to strawberry/potato. Load durations and numerical increments are solver settings, not damage tolerances.}
    \label{tab:app_solver_config}
    \scriptsize
    \setlength{\tabcolsep}{3pt}
    \renewcommand{\arraystretch}{1.18}
    \begin{tabular}{@{}p{0.17\linewidth}p{0.255\linewidth}p{0.255\linewidth}p{0.255\linewidth}@{}}
        \toprule
        Setting & CalculiX 2.23 & FEBio 4.13 & OpenRadioss \\
        \midrule
        Integration & Implicit, large deformation & Implicit dynamic ramp & Explicit dynamic \\
        Object elements & S6 shells; C3D10 solids & Quadratic tetrahedra & DKT18 shells \\
        Shell thickness points & 3 (3 in-plane points) & Not used for the listed soft-solid cases & 5 \\
        Pad elements & Quadratic tetrahedra & Native deformable pad & TETRA10, LAW42 \\
        Contact & Surface-to-surface penalty & Auto-scaled penalty contact & TYPE24 finite sliding \\
        Contact parameter & $2\times10^{11}$\,Pa/m; friction 0.5 & Penalty scale 0.03; gap tolerance 5/10\,$\mu$m & Native contact stiffness; friction inherited from input \\
        Loading & Prescribed backing displacement & Prescribed backing displacement & Prescribed backing displacement \\
        Time control & Automatic static increments & Adaptive time-step controller & Explicit stability-controlled step \\
        Ramp profile & Piecewise prescribed history & Ramp scale 5/1\,s; assessed fraction 0.5 & Case-specific; 2\,oz: 0.08\,s, displacement factor 4 \\
        Increment settings & At most 10,000 increments per stage; minimum $10^{-10}$ & Maximum ramp step 0.05/0.02\,s; initial cap 0.01/0.002\,s & Stable step recorded with each input deck \\
        Nonlinear settings & Automatic cutback & $r_{\rm tol}=10^{-10}$, $e_{\rm tol}=10^{-3}$, $d_{\rm tol}=0$; 12 retries; 100 target iterations & Energy and force-balance checks on the decision interval \\
        Linear solve / device & cuDSS on GPU; host routines on CPU & cuDSS plugin on GPU; host routines on CPU & CPU explicit route \\
        \bottomrule
    \end{tabular}
\end{table}

The implicit static route adapts its increments as contact and material nonlinearities evolve. For dynamic routes, a gradual displacement ramp reduces transient inertial effects; slower ramps are used when the decision interval fails the prescribed numerical checks. The FEBio ramp scale and its evaluated fraction jointly determine the loaded duration, while the OpenRadioss displacement factor changes the imposed closure amplitude. These parameters must therefore be distinguished from an allowable plastic-strain or damaged-volume fraction. Numerical checks are applied to the interval supporting the damage decision, with the object-specific criterion determining the relevant interval.

\subsection{Damage Thresholds and Servo Coupling}
\label{app:fem_thresholds}

Along the displacement-controlled loading path, we record the pad reaction together with object stresses, strains, and the applicable damage indicator. Let $q$ denote the total additional closure of both pad backings from initial contact, including pad and object compression, and let $R_k(q)$ denote the per-jaw reaction for grasp $g_k$. At the first criterion crossing $q_d$, the FEM damage load is $R_k(q_d)$.

To account for the decrease in position error as the pads and object compress, we use the recorded contact force $f_k$ as the undeformed reference load. Under a symmetric, quasi-static, unsaturated approximation with a fixed grasp direction, the additional closure reduces the servo load by
\begin{equation}
    \Delta F_{\mathrm{servo}}(q)=\kappa q,
    \qquad \kappa=\frac{k_p}{2}.
    \label{eq:app_servo_load_reduction}
\end{equation}
The factor of two follows from using total closure with a per-jaw force convention. Thus $k_p=1000$\,N/m gives $\kappa=500$\,N/m. This is an incremental approximation around the recorded force reference, using the nominal controller stiffness.

On a stable loading branch reachable from initial contact, equilibrium satisfies $R_k(q)=f_k-\kappa q$. Expressing the criterion crossing in the recorded-force reference gives the threshold used in Sec.~\ref{sec:solver_based_assessment}:
\begin{equation}
    F_{\mathrm{dmg}}(g_k)=R_k(q_d)+\kappa q_d.
    \label{eq:app_servo_damage_threshold}
\end{equation}
Here $R_k(q_d)$ remains the physical damage load under pad contact; $F_{\mathrm{dmg}}(g_k)$ is its equivalent comparison threshold. Material properties and damage criteria are unchanged. In the notation of Sec.~\ref{sec:solver_based_assessment}, the solver procedure comprises this FEM solve together with its servo-equivalent conversion. The controller gain enters only through $\kappa$ and is fixed across grasps within a platform, so it is suppressed in that argument list along with the pad geometry and the boundary-condition convention.

For nonmonotonic responses, including buckling, the conversion requires a reachable stable branch. Actuator saturation requires the corresponding bounded controller relation. A crossing between saved states retains a threshold interval, and later force reduction does not undo an earlier damage event.

When a FEM analysis fails to provide a usable damage threshold due to nonconvergence, we estimate the threshold using the median of the directly computed thresholds from the five nearest grasps of the same object. Proximity is determined by the recorded contact-point locations. The estimated threshold is compared with the target grasp's recorded load using the same force convention. Only valid, directly computed FEM thresholds are used as references.

\subsection{Computation and GPU Acceleration}
\label{app:fem_cost}

The assessment is performed offline, after policy execution, and its cost is associated with the grasp-specific contact problem rather than with each policy inference. We cache computed curves for reuse with matching geometry, material, contact, and solver configurations. When the saved force--displacement and damage histories cover the required loading path, servo coupling is evaluated by postprocessing these curves without an additional FEM solve. GPU acceleration in CalculiX and FEBio targets the sparse linear-system solve through cuDSS; constitutive updates, element and contact assembly, and nonlinear iterations remain in the host solvers. The adopted OpenRadioss LAW36/DKT18 route runs on CPUs.

In the reported evaluations, FEBio runs have median wall times of 273\,s for the boiled potato and 531\,s for the strawberry per attempted grasp configuration. A representative three-thread OpenRadioss 2\,oz-cup solve takes 49.48\,min. These timings describe their respective recorded configurations, including unsuccessful attempts in the FEBio batch statistics; they are not a controlled comparison between solvers. Parallel execution and cache reuse reduce the total elapsed time without changing the underlying material and contact models.

\paragraph{Scope of the offline assessment.}
The contact problems above are constructed from grasps recorded in the host rigid-body simulator. The FEM contact model accounts for changes in pressure and stress distributions, and the offline servo coupling accounts for load reduction due to pad and object compression under the stated approximation. The recorded grasp configuration and policy actions are held fixed: deformed geometry is not fed back as a new policy observation or used to rerun the task dynamics. The result therefore assesses object preservation conditional on the recorded grasps and the specified mechanical and controller models.

\clearpage

\section{Object-Specific Relabeling and Training}
\label{app:relabel_training}

\subsection{Instruction-Based Episode Selection}
\label{app:relabel_selection}

\begin{figure}[t]
    \centering
    \includegraphics[width=\linewidth]{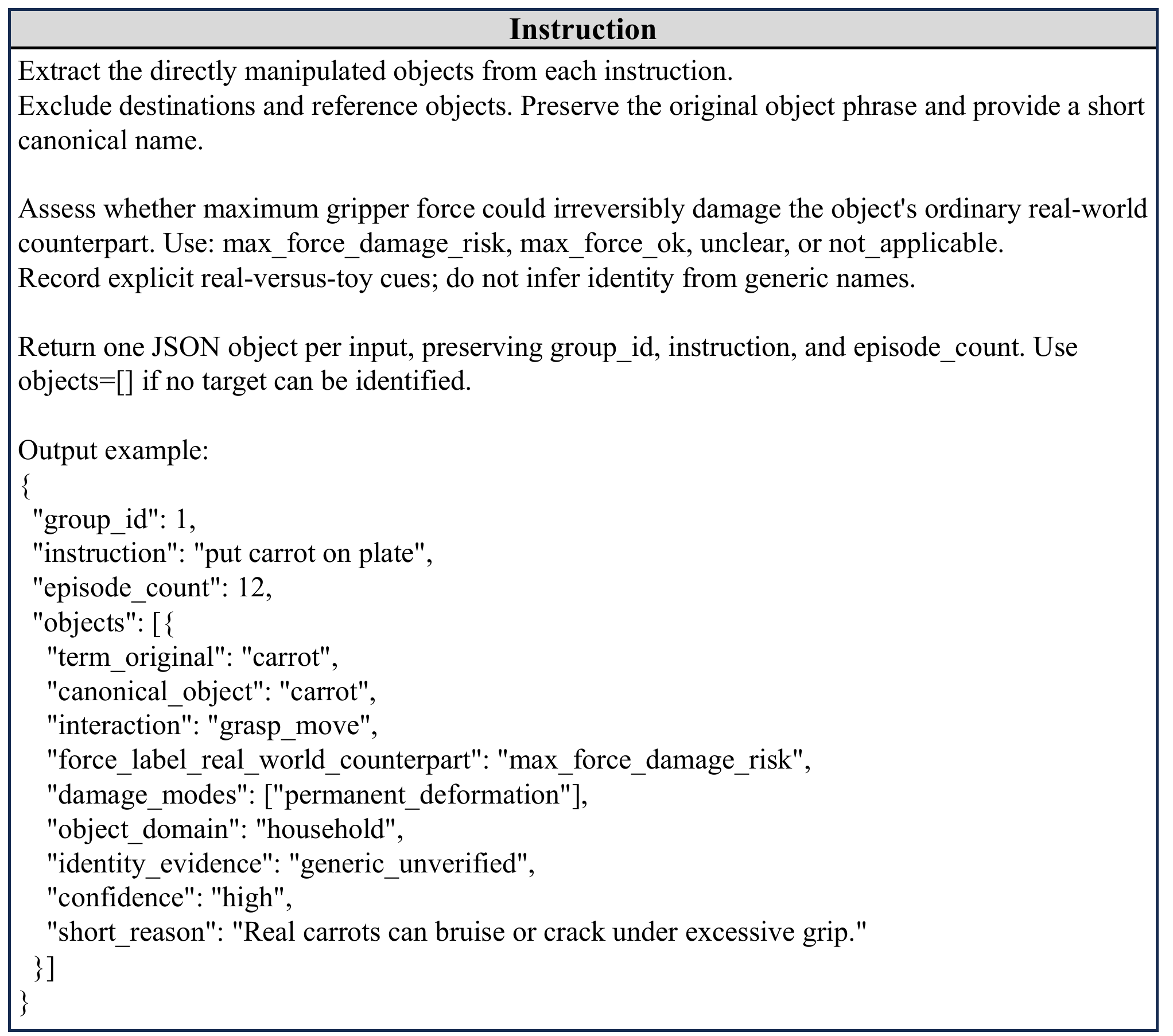}
    \caption{Object-extraction and Damage-risk annotation instructions used for Bridge demonstrations.}
    \label{fig:app_object_prompt}
\end{figure}

We retain the full eligible Bridge training mixture~\citep{bridgedata_v2} and replace the closing-gripper labels of selected demonstrations. Of 53,192 episodes, 38,660 have nonempty instructions, corresponding to 19,973 distinct instruction groups. We annotate these groups using GPT-5.6-Luna~\citep{openai2026gpt56} via the API with low reasoning effort, then map the annotations back to their episodes. The annotation prompt is shown in Fig.~\ref{fig:app_object_prompt}.

The annotation preserves the original object phrase and assigns a canonical name, an interaction type, and a damage-risk label. It targets the directly manipulated object, excluding destinations and reference objects. The risk label concerns irreversible damage to the named object's ordinary real-world counterpart; explicit real-versus-replica cues are recorded separately rather than inferred from a generic food name.

We consolidate the names with an explicit alias dictionary and deterministic string rules, yielding 170 concepts from 679 raw canonical strings, including unresolved concepts. The procedure standardizes spelling, merges listed aliases, strips supported modifiers, and removes non-object fragments. A subsequent filter checks the manipulation phrase and grasp interaction before assigning an episode to a force target (Table~\ref{tab:app_normalization}).

\begin{table}[t]
    \centering
    \caption{Rule-based consolidation and target-object filtering. The saved mapping records each name's disposition and its reason.}
    \label{tab:app_normalization}
    \small
    \renewcommand{\arraystretch}{1.13}
    \begin{tabular}{@{}p{0.26\linewidth}p{0.68\linewidth}@{}}
        \toprule
        Stage & Operation \\
        \midrule
        Standardization & Lowercase, normalize whitespace, and replace underscores/hyphens with spaces. \\
        Alias mapping & Merge registered variants, e.g., brinjal $\rightarrow$ eggplant and maize/corncob $\rightarrow$ corn. \\
        Modifier cleanup & Remove supported leading modifiers and explicit spatial/state suffixes before alias matching. \\
        Extraction cleanup & Remove pronouns, relation fragments, and malformed labels; preserve unresolved object-like names separately. \\
        Aggregation & Combine identical canonical categories using the saved mapping. \\
        Episode filtering & Require a policy-covered object to be the directly manipulated target of a grasp interaction. \\
        \bottomrule
    \end{tabular}
\end{table}

The force-target mapping covers 74 categories: 50 damage-sensitive categories and 24 additional container, packaging, or related categories. Their annotation-based candidate set contains 8,989 distinct episodes; target filtering retains 8,836 episodes (16.61\% of the dataset). The remaining 44,356 episodes retain their original labels. Table~\ref{tab:app_relabel_counts} distinguishes the instruction, category, and episode counts.

The existing training index contains 51,019 eligible episodes and 1,803,482 frames. All 8,836 selected episodes are included in this index; the other 42,183 training episodes retain their labels. Selection therefore changes supervision within the full training mixture rather than restricting training to the selected subset.

\begin{table}[t]
    \centering
    \caption{Data selection for object-specific relabeling.}
    \label{tab:app_relabel_counts}
    \small
    \begin{tabular}{@{}l r l@{}}
        \toprule
        Stage & Count & Unit \\
        \midrule
        Full Bridge dataset & 53,192 & Episodes \\
        Nonempty instructions & 38,660 & Episodes \\
        Distinct nonempty instructions & 19,973 & Instruction groups \\
        Raw canonical names & 679 & Strings \\
        Consolidated concepts & 170 & Categories, including unresolved \\
        Force-policy categories & 74 & Object categories \\
        Candidate episode union & 8,989 & Episodes \\
        Selected for relabeling & 8,836 & Episodes (16.61\%) \\
        Original labels retained (source index) & 44,356 & Episodes \\
        Eligible training index & 51,019 & Episodes \\
        Original labels retained (training index) & 42,183 & Episodes \\
        \bottomrule
    \end{tabular}
\end{table}

\subsection{Object-Conditioned Force Targets}
\label{app:relabel_force}

We use material and load information in \proposedone{}, available literature, and related-object proxies to determine the force targets. When both a lifting requirement $F_{\mathrm{lift},o}$ and a damage-load estimate $\widehat F_{\mathrm{dmg},o}$ are available, with $F_{\mathrm{lift},o}<\widehat F_{\mathrm{dmg},o}$, we set the target force to their midpoint:
\begin{equation}
  F_o^{\mathrm{target}}=\frac{F_{\mathrm{lift},o}+\widehat F_{\mathrm{dmg},o}}{2}.
  \label{eq:app_force_midpoint}
\end{equation}
For categories lacking a damage-load estimate, we use representative size and mass assumptions to estimate the lifting requirement and add a grip margin. The complete category-to-target mapping and its supporting evidence are provided with the implementation. In this experiment, the targets are implemented through category-level load proxies and a shared force cap. 

For approximate lifting feasibility, a symmetric grasp with per-jaw normal force $F$ supports mass $m$ when $2\mu F\gtrsim mg$. We use $\mu=0.5$ for this check. The masses describe representative objects rather than measurements of every recorded instance, and the resulting supervision proxies do not guarantee lifting or preservation for every instance.

\subsection{Force-to-Action Conversion}
\label{app:relabel_action}

Let $F_o$ be the nominal per-jaw force target for object category $o$. We use the position-servo relation in App.~\ref{app:gripper_nominal} to obtain its closing offset:
\begin{equation}
    \delta_o=\frac{F_o}{k_p},\qquad k_p=1000\,\mathrm{N/m}.
    \label{eq:app_force_offset}
\end{equation}
Thus $\delta_o$ in millimeters equals the numerical value of $F_o$ in newtons. This is a nominal force-to-position conversion, not a measured contact-force calibration; assessment still uses the contact loads recorded during evaluation.

For each closing interval $r$, we shift its observed-state window by one step, clip it at the episode boundary, and take the median of its second half, denoted by $\tilde g_r$. The recorded label convention is zero for closed and one for open. The achieved per-jaw position of the WidowX gripper in meters is
\begin{equation}
    q^{\mathrm{ach}}_r=0.010+0.025\tilde g_r.
    \label{eq:app_achieved_position}
\end{equation}
For selected episodes $\mathcal E_{\mathrm{sel}}$, only closing labels ($g_t<0.5$) are replaced:
\begin{equation}
    g'_t=
    \begin{cases}
        \operatorname{clip}_{[0,1]}
        \left(\dfrac{q^{\mathrm{ach}}_r-\delta_o-0.014}{0.024}\right),
        & e\in\mathcal E_{\mathrm{sel}},\ g_t<0.5,\\[6pt]
        g_t, & \text{otherwise}.
    \end{cases}
    \label{eq:app_relabel_action}
\end{equation}
Here $\delta_o$ is expressed in meters. Arm actions, opening commands, and unselected episodes are unchanged. Relabeling occurs in the data loader without rewriting the source dataset. This label convention is the reverse of the executed-closure coordinate in App.~\ref{app:gripper_distribution}.

\subsection{Training Configuration}
\label{app:relabel_optimization}

We train QwenGR00T~\citep{starvla} on the full eligible Bridge mixture with the selected closing labels replaced. Its vision-language component is initialized from Qwen3-VL-4B-Instruct~\citep{qwen3vl}, and its action model is initialized anew. We apply LoRA to the language component's attention and feed-forward projections, freeze the remaining vision-language parameters, and train the full action model. Table~\ref{tab:app_relabel_training} gives the configuration used for the reported model.

\begin{table}[t]
    \centering
    \caption{QwenGR00T object-specific supervision configuration. Sample exposure counts sampled training instances, including repetitions, rather than distinct demonstrations.}
    \label{tab:app_relabel_training}
    \footnotesize
    \renewcommand{\arraystretch}{1.12}
    \begin{tabular}{@{}p{0.39\linewidth}p{0.56\linewidth}@{}}
        \toprule
        Setting & Value \\
        \midrule
        Initialization & Qwen3-VL-4B-Instruct + new action model \\
        Training data & Bridge; 51,019 eligible episodes \\
        Trainable modules & Language LoRA + full action model \\
        LoRA rank / scale / dropout & $32 / 64 / 0.05$ \\
        LoRA targets & q/k/v/o and gate/up/down projections \\
        Trainable parameters & 227.533\,M \\
        Microbatch per GPU / GPUs / accumulation & $12 / 7 / 3$ \\
        Effective batch size & 252 \\
        Evaluated checkpoint & 20,000 optimizer updates \\
        Sample exposure at evaluation & 5,040,000 \\
        Learning rates & VLM LoRA $10^{-5}$; action model $10^{-4}$ \\
        Optimizer & AdamW; $\beta=(0.9,0.95)$; $\epsilon=10^{-8}$ \\
        Weight decay / gradient clipping & $10^{-8}$ / norm 1 \\
        Schedule & Cosine; minimum LR $5\times10^{-7}$ \\
        Scheduler horizon / warmup & 100,000 / 10,000 updates \\
        Precision / seed & BF16 / 42 \\
        Action dimension / horizon & 7 / 16 \\
        Diffusion repeats per training sample & 4 \\
        Checkpoint interval & 5,000 updates \\
        \bottomrule
    \end{tabular}
\end{table}

The run completes 20,318 optimizer updates, corresponding to 5,120,136 sampled training instances; the evaluation uses the 20,000-update checkpoint. The learning-rate schedule retains its 100,000-update horizon and is not compressed to the stopping budget. LoRA weights are merged for evaluation. This configuration uses Bridge-only data and VLM LoRA, whereas the public checkpoint uses a Bridge--RT-1 mixture and a different training configuration; the comparison therefore evaluates this supervision setting rather than isolating label replacement under otherwise identical training conditions.

\clearpage

\section{Gripper Command Distributions}
\label{app:gripper_distribution}

\subsection{Executed Closure and Aggregation}
\label{app:gripper_aggregation}

We analyze the joint targets actually sent to the simulator. Repeated physics-step records are reduced to one command per policy control step, and joint targets are converted to a shared closure coordinate: zero is fully open and one is fully closed. For left and right joint targets $q_L,q_R$ in meters,
\begin{equation}
    c_{\mathrm{WidowX}}=
    \frac{0.038-(q_L+q_R)/2}{0.024},
    \qquad
    c_{\mathrm{Panda}}=1-\frac{|q_L-q_R|}{0.08}.
    \label{eq:app_executed_closure}
\end{equation}

We include all policy control steps, including unsuccessful episodes and non-contact intervals. Let $T_{moe}$ be the number of control steps for model $m$, object $o$, and episode $e$. For bin $B_b$, the episode histogram and model--object average are
\begin{equation}
    p_{moe,b}=\frac{1}{T_{moe}}\sum_{t=1}^{T_{moe}}
       \mathbf{1}[c_{moet}\in B_b],
    \qquad
    p_{mo,b}=\frac{1}{E_{mo}}\sum_{e=1}^{E_{mo}}p_{moe,b}.
    \label{eq:app_gripper_histogram}
\end{equation}
Object-level figures average $p_{mo,b}$ equally over the included models, preventing longer episodes from dominating. Stored statistics use 20 equal-width bins; compact figures combine adjacent pairs into 10 bins. The vertical values are probability mass per displayed bin, although the word \emph{fraction} is omitted from plot axes.

For the contact-conditioned mean, we retain a control step if at least one recorded sample has simultaneous left- and right-jaw contact with the object, using a numerical tolerance of $10^{-8}$\,N on each recorded contact force. We average closure over these steps within each episode and then equally across episodes with recorded bilateral contact, including task failures. Episodes without bilateral contact do not contribute to this conditional mean; an object with no such observations has no contact-conditioned mean.

\subsection{Model-Wise Distributions Across Physics Axes}
\label{app:gripper_additional}

\begin{figure}[p]
    \centering
    \includegraphics[width=\linewidth]{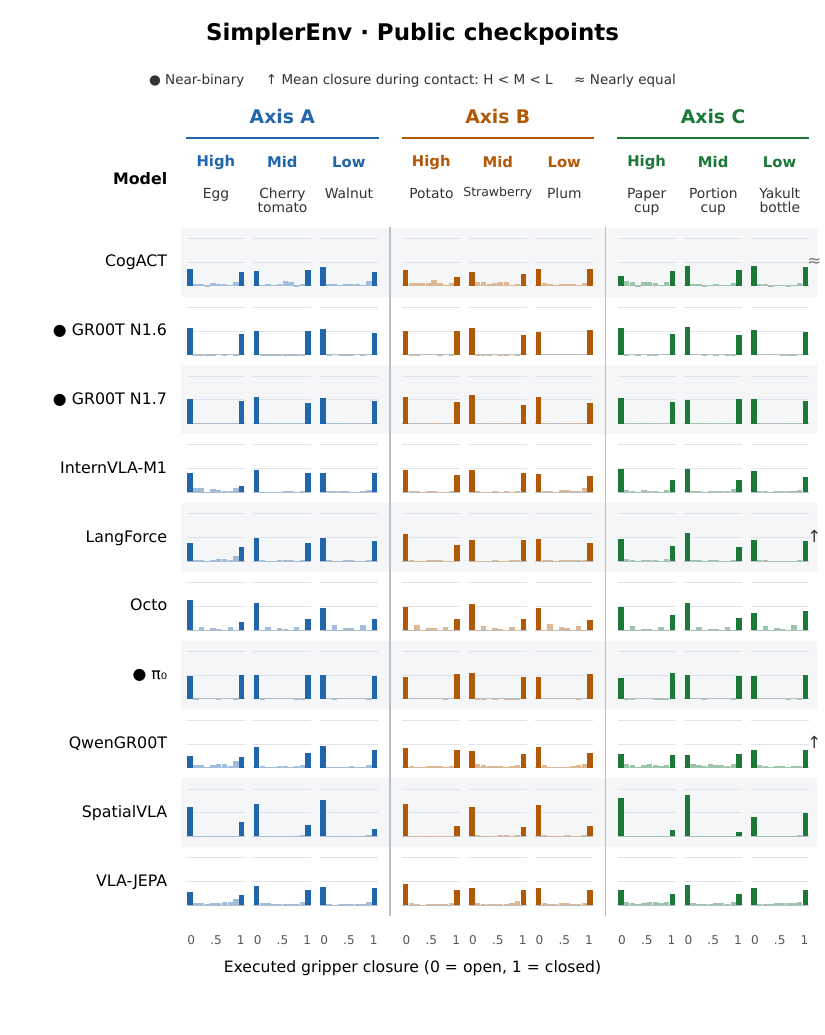}
    \caption{Gripper command distributions for all evaluated public checkpoints on SimplerEnv. Histograms and near-binary markers use all rollout steps; arrows and approximate-equality markers use mean closure during bilateral object contact. Each histogram has a common vertical range of 0--1 probability mass per bin. Marker definitions and aggregation are given in App.~\ref{app:gripper_aggregation} and App.~\ref{app:gripper_additional}.}
    \label{fig:app_gripper_models_simpler}
\end{figure}

\begin{figure}[p]
    \centering
    \includegraphics[width=\linewidth]{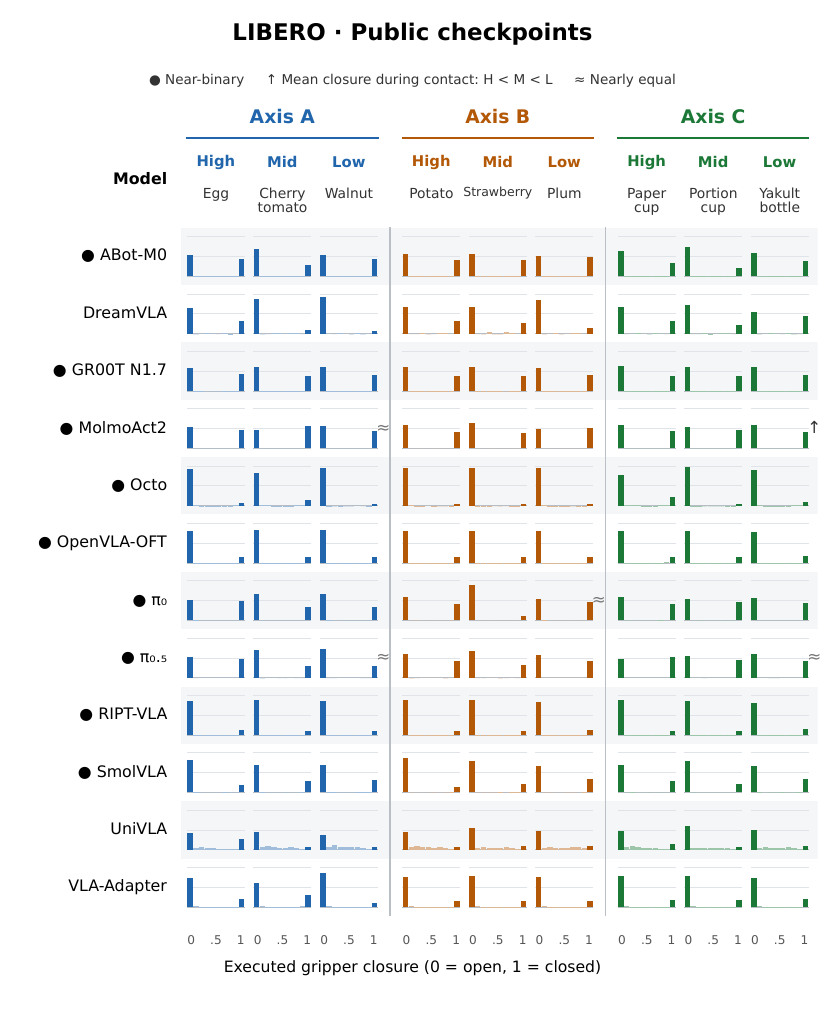}
    \caption{Gripper command distributions for all evaluated public checkpoints on LIBERO, with the same histogram scales and marker definitions as Fig.~\ref{fig:app_gripper_models_simpler}. An absent ordering marker can indicate either that the display conditions are not met or that an object has no recorded bilateral contact; missing contact is not counted as an ordering failure.}
    \label{fig:app_gripper_models_libero}
\end{figure}

Figures~\ref{fig:app_gripper_models_simpler} and~\ref{fig:app_gripper_models_libero} show all-rollout distributions for the public checkpoints on SimplerEnv and LIBERO, respectively. Each row is a model, and the nine columns retain the High/Mid/Low objects separately within each axis. Each model--object histogram averages 12 episodes equally. The figures therefore expose model-specific variation that can be obscured by the model-averaged distribution in Fig.~\ref{fig:gripper_commands}.

The markers summarize two distinct scopes. A filled circle before a model name indicates that, for each of its nine objects, at least 95\% of the all-rollout distribution lies in $c<0.1$ or $c\geq0.9$. The arrow and approximate-equality markers instead use the contact-conditioned means $\mu_H,\mu_M,\mu_L$: $\uparrow$ denotes $\mu_M-\mu_H>0.01$ and $\mu_L-\mu_M>0.01$, while $\approx$ denotes a range of at most 0.01 across the three means. No ordering marker is assigned when an object lacks bilateral-contact observations. These are descriptive display rules, not statistical significance tests or physical safety criteria.

\clearpage

\section{Difficulty Levels}
\label{sec:appendix_difficulty}

Each object can be lifted without damage only within a range of gripping forces. We characterize this range using the minimum lifting force $F_{\mathrm{lift}}$, the pad--object damage reaction $R_{\mathrm{dmg}}(g)$, and a nominal reference ceiling $F_{\mathrm{ceil}}(g)=\min\{k_p w(g)/2,F_{\mathrm{ref}}\}$. Here $w(g)$ is the contact-point distance for grasp $g$, $k_p=1000$\,N/m, and $F_{\mathrm{ref}}=20$\,N. The lifting force follows the reference asset's mass, friction, and center-of-mass offset. This index uses the mechanical reaction at damage, before the servo-equivalent threshold conversion in App.~\ref{app:fem_thresholds}; its reference ceiling is not a measured maximum contact force.
\begin{equation}
    D = 1 - \mathbb{E}_{g}\!\left[
    \frac{\ln\!\bigl(\min\{R_{\mathrm{dmg}}(g),\,F_{\mathrm{ceil}}(g)\}/F_{\mathrm{lift}}\bigr)}
         {\ln\!\bigl(F_{\mathrm{ceil}}(g)/F_{\mathrm{lift}}\bigr)}
    \right].
    \label{eq:difficulty}
\end{equation}
The term in brackets is set to zero when $F_{\mathrm{ceil}}(g)\leq F_{\mathrm{lift}}$ or $R_{\mathrm{dmg}}(g)\leq F_{\mathrm{lift}}$, giving $D\in[0,1]$. We compute the logarithmic safe-force fraction for each grasp and then average it. Thus $D=0$ means that the assessed thresholds are not reached within the reference range, while larger values indicate a narrower lifting range without damage. This is a physical reference index, not a task-success statistic or a claim of platform-independent contact forces.

Table~\ref{tab:difficulty_values} summarizes the available grasp-specific FEM evidence. We deduplicate grasp records by their cache keys and use thresholds with an observed criterion crossing and a corresponding contact width; right-censored or invalid curves and inferred thresholds are excluded from this summary. Consequently, the medians and $D$ values describe this assessed subset. $D$ is averaged over individual grasps, not computed from the displayed median. The two paper-cup variants are listed separately; the benchmark's High/Medium/Low assignments are retained.

\begin{table}[t]
    \caption{Reference lifting force, median pad--object damage reaction, and difficulty over $n$ bracketed grasps. Forces are per jaw, in newtons. S and L denote the SimplerEnv and LIBERO paper-cup variants. Other rows pool available grasps from both platforms.}
    \label{tab:difficulty_values}
    \centering
    \small
    \begin{tabular}{@{}llrrrrl@{}}
        \toprule
        Axis & Object & $F_{\mathrm{lift}}$ & Median $R_{\mathrm{dmg}}$ & $n$ & $D$ & Level \\
        \midrule
        A & Egg & 0.425 & 5.05 & 285 & 0.357 & High \\
        A & Cherry tomato & 0.568 & 11.01 & 236 & 0.185 & Medium \\
        A & Walnut & 0.098 & 63.50 & 45 & 0.000 & Low \\
        \addlinespace
        B & Boiled potato & 0.627 & 8.15 & 212 & 0.242 & High \\
        B & Strawberry & 0.152 & 6.40 & 72 & 0.147 & Medium \\
        B & Plum & 0.519 & 32.34 & 11 & 0.001 & Low \\
        \addlinespace
        C & Paper cup (S) & 0.011 & 1.79 & 55 & 0.326 & High \\
        C & Paper cup (L) & 0.573 & 3.20 & 26 & 0.521 & High \\
        C & Portion cup & 0.012 & 1.60 & 33 & 0.302 & Medium \\
        C & Yakult bottle & 0.396 & 16.96 & 83 & 0.052 & Low \\
        \bottomrule
    \end{tabular}
\end{table}

\clearpage
\section{Low-Friction Ice Cube Evaluation}
\label{app:ice_gripping}

\begin{figure}[ht]
    \centering
    \includegraphics[width=0.65\linewidth]{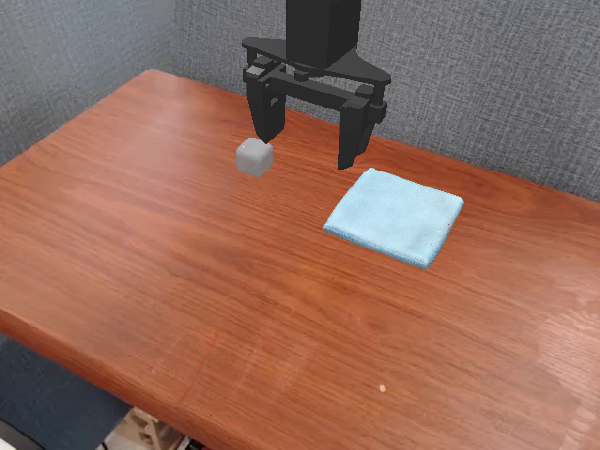}
    \caption{Initial rollout view of the ice cube object.}
    \label{fig:app_ice_asset}
\end{figure}

We use the same SimplerEnv put-on-towel task with an ice cube object, shown in Fig.~\ref{fig:app_ice_asset}. We evaluate the public and retrained QwenGR00T models over the same 12 initial configurations and report SR using the platform's task-success criterion. The representative friction coefficient is set to $\mu=0.12$ before evaluation, based on the low-friction regime reported for rubber--ice contact~\citep{tada2023rubberice}; this is a literature-informed proxy for the pad contact. Static and dynamic friction are both set to this value on the object and jaw surfaces. Mean closure in Table~\ref{tab:ice_gripping} is averaged over bilateral-contact steps within each episode and then across episodes with such contact.

\end{document}